\documentclass[runningheads]{llncs}

\usepackage{eccv}

\usepackage{eccvabbrv}

\usepackage{graphicx}
\usepackage{booktabs}

\usepackage{multirow}
\usepackage{float}

\usepackage[accsupp]{axessibility}  

\usepackage{hyperref}

\usepackage{orcidlink}

\begin{document}

\title{Cross-Spectral Dense Correspondence for Multimodal Spectral Medical Imaging} 

\titlerunning{Cross-Spectral Dense Correspondence}

\author{Eric L. Wisotzky\inst{1,2}\orcidlink{0000-0001-5731-7058} \and
Jost Triller\inst{1} \and
Simon W. Härtl \inst{3}\orcidlink{0009-0007-3843-5850} \and
Oliver T. Bruns \inst{3}\orcidlink{0000-0002-5738-0126} \and
Peter Eisert \inst{1,2}\orcidlink{0000-0001-8378-4805} \and
Anna Hilsmann \inst{1}\orcidlink{0000-0002-2086-0951}}

\authorrunning{E.~L.~Wisotzky et al.}

\institute{Vision \& Imaging Technologies, Fraunhofer Heinrich Hertz Institute HHI, Berlin, Germany\\
\email{\{name.surname\}@hhi.fraunhofer.de} \and
Visual Computing, Humboldt University Berlin, Germany \and 
Functional Imaging in Surgical Oncology, National Center for Tumordiseases, Dresden, Germany}

\maketitle

\begin{abstract}
Precise dense correspondence is a fundamental prerequisite for multimodal spectral imaging systems that fuse disparate wavelength ranges for subsequent analysis in medical and scientific imaging.
Corresponding image points are often observed with non-overlapping spectral sensitivities, leading to wavelength-dependent contrast changes, intensity inversions, and appearance shifts for which dense ground truth is difficult to obtain and conventional RGB-based training data provides only limited supervision.
We address this data gap by introducing a sensor-agnostic cross-spectral modulation protocol on established correspondence benchmarks with intensity input projection, and by proposing a synthetic cross-spectral correspondence benchmark simulating physically plausible radiometric differences.
Evaluation on several modern dense correspondence backbones trained with our unified cross-spectral protocol showed substantial improvements under severe spectral mismatch while maintaining performance on standard RGB benchmarks.
Ablation experiments show that view-dependent channel selection and nonlinear radiometric transformations provide complementary robustness, indicating that the primary limitation of existing models is not their structural matching capacity but the mismatch between training distribution and spectral characteristics of the target image pair.
Qualitative evaluations on heterogeneous medical spectral acquisition systems demonstrate the practical relevance of the proposed training data augmentation protocol as an enabler for spatially coherent spectral fusion in HSI workflows.
\end{abstract}

\section{Introduction} \label{sec:intro}
Capturing scene appearance beyond the visible spectrum enables imaging systems to access material properties and physiological cues that are weak or ambiguous in conventional RGB images. This is particularly relevant for multi-sensor hyperspectral imaging (HSI) systems, where complementary wavelength ranges can support spatially resolved tissue assessment, spectral fusion, and, where applicable, depth recovery for image-guided surgery. The practical use of such systems depends on accurate spatial alignment across views, optical paths, and spectral channels. Before spectral measurements can be fused into a coherent representation, observations acquired with different sensors or wavelength sensitivities must be brought into accurate dense pixel-wise correspondence.

A central challenge in this setting is that corresponding image points may be acquired with non-overlapping spectral sensitivities and sensor-specific radiometric responses. As a result, cross-view intensities are not photometrically consistent and may exhibit wavelength-dependent contrast suppression, contrast enhancement, or even intensity inversion. These effects violate the brightness-constancy assumptions underlying classical stereo and many modern correspondence pipelines. At the same time, dense ground-truth correspondence data are difficult to obtain for real surgical HSI, since tissue motion, specularities, moist surfaces, limited texture, and clinical acquisition constraints make precise annotation or calibration-based ground truth challenging.

Robust intraoperative spectral fusion is therefore limited not only by model capacity, but also by the absence of curated training and evaluation data that represent realistic cross-spectral appearance shifts. Standard dense correspondence benchmarks are predominantly based on RGB imagery, where appearance differences between views are comparatively moderate and do not reflect the material-dependent radiometric changes observed across distant wavelength ranges. 
To address this gap, we introduce a unified data generation framework for cross-spectral dense correspondence. Starting from established RGB correspondence benchmarks, it generates cross-spectral training and evaluation data by simulating wavelength-dependent radiometric responses while preserving the original dense ground truth. We further complement these data with a synthetic cross-spectral benchmark that enables controlled evaluation under realistic wavelength-dependent appearance variation.

Building on these data, our focus is not on designing a task-specific correspondence architecture, but on developing a sensor-agnostic training and evaluation framework that enables existing correspondence models to generalize across heterogeneous multi- and hyperspectral imaging systems.
To this end, we combine a sensor-agnostic input representation with spectral-response modulation during training, explicitly exposing the correspondence model to wavelength-dependent radiometric variations and contrast inversions. By learning invariance to these modality-induced appearance changes, the resulting framework establishes reliable dense correspondences even when classical brightness-constancy assumptions are fundamentally violated. These dense alignments provide the foundation for spatially coherent spectral fusion and, where applicable, depth recovery across heterogeneous spectral acquisition systems.

We evaluate the proposed framework quantitatively on established benchmarks and cross-spectral variants to assess accuracy and robustness under controlled spectral mismatch. In addition, we introduce a synthetic cross-spectral dataset that enables systematic evaluation under non-overlapping spectral sensitivities.
Furthermore, we demonstrate the generality of our framework across three representative cross-spectral medical settings for which dense ground truth is unavailable, covering a broad range of geometric configurations, spectral separations, and radiometric characteristics encountered in practice.

In summary, we make the following contributions:
\begin{itemize}
    \item \textbf{Cross-spectral augmentation:} We introduce a sensor-agnostic training protocol that models wavelength-dependent radiometric variation, including contrast suppression and intensity inversion, for robust dense correspondence under spectral mismatch.
    \item \textbf{Synthetic cross-spectral benchmark:} We provide a synthetic cross-spec\-tral dataset with dense ground-truth displacement and controlled wavelength-pair variation to systematically evaluate correspondence under non-overlap\-ping spectral sensitivities, see \url{github.com/fraunhoferhhi/XSpecCorr}
    \item \textbf{Architecture-level analysis:} We evaluate multiple modern dense correspondence models and show that the proposed training strategy consistently improves robustness across architectures, demonstrating that compatibility between the training distribution and the target spectral domain is a dominant factor for robust cross-spectral correspondence.
    \item \textbf{Clinical relevance:} We demonstrate applicability of the approach on heterogeneous cross-spectral acquisition settings and discuss its role as an enabling component for spatially coherent spectral fusion in stereo-HSI and HSI light-field workflows.
\end{itemize}

\section{Related Work} \label{sec:relatedwork}
Medical MSI and HSI extend conventional RGB imaging by acquiring spatially resolved spectral signatures that can encode material composition, tissue physiology, and optical absorption properties \cite{lu2014medical}. This has motivated applications in surgical scene understanding, tissue characterization, perfusion assessment, and spectral reconstruction for image-guided interventions. Recent work on surgical HSI has shown its potential for organ segmentation, domain-robust surgical scene analysis, and reconstruction of spectral signatures \cite{seidlitz2022robust,wisotzky2024automatic}. Practical spectral imaging systems acquire wavelength information through sequential scanning \cite{clancy2021intraoperative,wisotzky2018intraoperative}, snapshot mosaics \cite{wisotzky2019validation}, multi-camera rigs \cite{wisotzky2020validation, wisotzky2025continuous}, or spectral light-field designs \cite{muhle2021comparison,maccormac2023lightfield, wisotzky20233d}. In such systems, spatial alignment becomes a prerequisite for spectral fusion \cite{wisotzky2024MSIfusion}, but corresponding image points may be observed under different viewpoints, optical paths, and non-overlapping spectral sensitivities. This makes correspondence estimation a central data and calibration problem.

In surgical imaging, synthetic data and targeted augmentation is highly relevant as annotated clinical data are limited and ground truth for geometry, correspondence, and tissue motion is difficult to obtain \textit{in-vivo}. Prior work has used simulations, phantoms, structured-light acquisition, or controlled \textit{ex-vivo} setups to generate reference data for reconstruction and scene understanding \cite{Butler:ECCV:2012, mayer2016large,azagra2023endomapper,scharstein2014high}. In surgical HSI, augmentation has been used to address geometric domain shifts, showing that data-centric modifications can substantially improve robustness without changing the underlying network architecture \cite{sellner2023semantic}. In contrast to generic image augmentation, cross-spectral correspondence requires augmentations that preserve geometric structure while changing the radiometric relationship between paired views. Our work follows this data-centric perspective and focuses on curating spectral-response augmentations and synthetic evaluation data that expose correspondence models to wavelength-dependent appearance changes.

Dense correspondence and camera calibration are fundamental components of surgical 3D perception, augmented reality, and image-guided intervention \cite{wisotzky2025telepresence}. 
Public stereo-endoscopic datasets and challenges have enabled quantitative evaluation of reconstruction methods \cite{azagra2023endomapper}.
Nevertheless, surgical acquisition remains challenging due to weak texture, specular reflections, smoke, blood, tissue deformation, breathing motion, limited calibration access, and workflow constraints. These difficulties are amplified in spectral imaging, where correspondence must be estimated not only across viewpoint changes but also across wavelength-dependent appearance changes. As a result, robust spectral fusion cannot rely solely on conventional stereo assumptions.

Modern dense correspondence architectures combine learned feature extraction, correlation reasoning, and iterative refinement, achieving strong performance under conventional RGB imaging conditions \cite{teed2020raft,lipson2021raft-stereo,xu2023unifying,liu2020flow2stereo}. These methods are trained on benchmarks where cross-view appearance variation remains moderate. Cross-spectral stereo and multi-modal matching address correspondence between images acquired in different spectral ranges, where brightness constancy can fail and local contrast may be suppressed, enhanced, or inverted \cite{zhi2018deep,brucker2024cross}. Existing approaches often focus on fixed modality pairs or rectified stereo configurations \cite{mehltretter2018multimodal,tuzcuouglu2024xoftr,huang2024data}, whereas practical spectral imaging systems may combine heterogeneous sensors \cite{tanriverdi2019dual,wisotzky2025real}, imperfect rectification \cite{park2026multimodal}, multiple spectral bands \cite{genser2020camera}, and light-field sub-views \cite{maccormac2023lightfield,kray2025intraoperative}. We therefore study cross-spectral correspondence as a data and augmentation problem. Instead of designing a task-specific matcher, we adapt existing high-capacity correspondence backbones through sensor-agnostic input representations, spectral-response modulation, and a synthetic dataset with dense ground-truth displacement.

\section{Modality-Robust Dense Correspondence} \label{sec:method}
In contrast to prior cross-spectral methods restricted to specific modality pairs and rectified geometries, we propose a unified dense two-dimensional (2D) correspondence framework that generalizes across heterogeneous spectral acquisition systems.
We formulate cross-spectral dense correspondence as the estimation of a pixel-wise displacement field between two images $I_a$ and $I_b$ of the same scene acquired from different viewpoints and with non-overlapping spectral sensitivities. The objective is to estimate a dense displacement field $\mathbf{u}(x,y) = (u(x,y), v(x,y))$ assigning each pixel in $I_a$ a corresponding location in $I_b$.

A central challenge arises from the heterogeneity of spectral imaging systems. MSI and HSI setups vary widely in band count, spectral coverage, and radiometric response functions, and in many multi-sensor configurations no shared spectral channel exists.
At the same time, the integrated cameras in such systems frequently cover disjoint wavelength ranges. As a result, corresponding regions may exhibit substantially different radiometric responses, and local contrast can be suppressed, distorted, or even inverted across views. This violation of photometric consistency makes naive similarity measures ambiguous and significantly complicates dense matching under spectral mismatch. Robust correspondence learning must therefore explicitly accommodate modality-induced radiometric variation. To ensure broad applicability across heterogeneous cameras, we design a data-centric cross-spectral correspondence training and data augmentation by spectral-response modulation.

Building on these principles, a modality-robust dense correspondence training framework for cross-spectral image pairs acquired under heterogeneous spectral sensitivities is proposed. The objective is to estimate a geometry-consistent pixel-wise displacement field while explicitly accounting for modality-induced appearance variation.
The proposed framework is built upon three design principles:
(i) a geometry-agnostic 2D displacement formulation,
(ii) a sensor-agnostic input representation that removes wavelength-specific channel semantics, and
(iii) explicit modeling of modality-induced radiometric variation during training.

Together, these principles explicitly separate spatial-structural alignment from wavelength-dependent appearance cues and decouple correspondence estimation from fixed spectral channel semantics. This structured cross-modality strategy enables generalization across heterogeneous spectral acquisition systems and is realized within iterative dense-matching architectures to provide a unified correspondence backbone.

\subsection{Training for Attaining Cross-Spectral Robustness}
\label{sec:method_network}
We realize the proposed cross-modality strategy within a dense correspondence framework that estimates full 2D displacement fields through feature extraction, correlation reasoning, and iterative refinement. The framework operates on paired images and predicts a displacement vector per pixel in the reference view. 
At its core, we use two training strategies:
(i) input representations are sensor-agnostic and independent of fixed wavelength semantics, and
(ii) radiometric variation induced by spectral mismatch is modeled explicitly during training.

To ensure applicability across heterogeneous MSI/HSI systems, we enforce a sensor-agnostic input model by projecting all inputs to a normalized single-channel intensity representation. Multi-band captures are normalized per band and represented by a single intensity channel, obtained either through intensity projection or by selecting an individual spectral band to prevent reliance on fixed wavelength cues. This representation removes channel semantics while preserving spatial layout, ensuring that the correspondence model operates consistently across heterogeneous spectral setups. This only requires a minor configuration of input interface for single-channel operation, while the remainder of the architecture remains unchanged.

\subsection{Data Augmentation by Spectral-Response Modulation} \label{sec:dataaugment}
To explicitly model modality-induced variation during training, we simulate spectral mismatch as a structured radiometric transformation applied to one view of each image pair. Instead of assuming photometric consistency between views, we introduce view-specific nonlinear intensity mappings that emulate sensor-dependent spectral responses. The applied transformations are structured and deterministic per training sample rather than stochastic pixel-wise perturbations. This ensures that spatial-structural consistency remains intact while the radiometric relationship between views varies in a controlled manner.

Starting from standard RGB correspondence datasets, we generate cross-spectral training pairs that mimic non-overlapping or strongly shifted spectral sensitivities while preserving the original spatial structure.
Each RGB image pair is converted into a cross-spectral pair using one of two modification strategies:
\begin{enumerate}
    \item \textbf{Non-overlapping channel pairing.} Different color channels are selected for the two views, e.g., blue for the left image and red for the right image, yielding intensity-like inputs derived from disjoint visible bands.
    \item \textbf{Grayscale with spectral response modulation.} We normalize and convert the RGB channels to grayscale, while one view undergoes a nonlinear radiometric transformation during grayscale conversion to emulate sensor-dependent spectral responses.
\end{enumerate}
The transformations are sampled from a fixed family of non-linear functions, including identity/inver\-sion, square-root variants, power laws ($n\in\{2,4\}$), and logarithmic mappings, see in the Supplementary Material. These transformations are an approximation of the dominant radiometric effects, explicitly inducing contrast suppression and inversion effects commonly observed in cross-spectral imagery, and individual transformations can be mapped to specific spectral tissue behaviors, e.g., in our RGB-SWIR data.
Both modification strategies yield single-channel inputs. 
Importantly, the globally applied nonlinear mappings complement the non-overlapping channel-pairing strategy: while the former introduces nonlinear contrast changes and inversions, the latter naturally produces spatially heterogeneous appearance changes as local structures exhibit different relative responses across RGB channels.

\section{Experiments}
To demonstrate generalizability and robustness of the proposed framework, we evaluate it using several established dense correspondence architectures. Representative examples include RAFT-style recurrent refinement and correlation-based matching \cite{teed2020raft}, as well as efficiency and robustness-oriented variants such as SKFlow \cite{sun2022skflow}, DIP \cite{zheng2022dip}, and SEA-RAFT \cite{wang2024sea}, and attention-augmented matching via global motion aggregation (GMA) \cite{jiang2021gma}.

We train and evaluate our framework on three complementary data sources:
(i) established dense correspondence benchmarks,
(ii) a synthetic cross-spectral dataset constructed to simulate controlled spectral mismatch, and
(iii) real-world medical data acquired with heterogeneous cross-spectral imaging systems.

\subsection{Public RGB Benchmark Datasets} \label{sec:method_benchprep}
To enable controlled training and evaluation of cross-spectral correspondence, we derive spectral-mismatch variants of established dense correspondence benchmark datasets (optical flow and stereo). In this work, we use the following datasets: \textit{FlyingChairs} \cite{dosovitskiy2015flownet}, \textit{FlyingThings3D} \cite{mayer2016large}, \textit{MPI-Sintel} \cite{Butler:ECCV:2012}, \textit{HD1K}\cite{kondermann2016hci}, and \textit{KITTI} \cite{Menze2015CVPR} for training and validation, as well as \textit{Middlebury Stereo} \cite{scharstein2014high}, \textit{ETH3D} \cite{schops2017multi}, and \textit{InStereo2K} \cite{bao2020instereo2k} for evaluation only.

\subsection{Synth: Controlled Benchmark for Spectral Mismatch} \label{sec:method_synth} 
Dense ground-truth correspondence is difficult to obtain for real surgical spectral imaging data, where tissue deformation, specularities, moist surfaces, limited texture, and clinical acquisition constrain capturing process. We therefore introduce \textit{Synth}, a synthetic cross-spectral correspondence benchmark designed to isolate two factors that are entangled in real acquisitions: geometric displacement and wavelength-dependent appearance change. However, \textit{Synth} is not a tissue simulator. It is a controlled benchmark for material-dependent cross-spectral radiometric shifts with exact dense geometry. The dataset contains procedurally generated triplets $(I_L, I_R, u')$ at $768 \times 512$ pixels, with dense ground-truth 2D displacement and 100 samples per motion regime and spectral range.

Instead of using arbitrary color transformations, the spectral appearance of each of the six layered polygonal objects is derived from measured reflectance spectra of the USGS Spectral Library \cite{kokaly2017usgs}. For each view, a virtual camera response is generated by selecting different wavelength ranges from the same material spectrum, with randomly perturbed band limits to emulate variability in spectral band placement and effective sensor response. This creates paired images that preserve object identity and scene geometry while inducing material-dependent cross-spectral contrast changes, including cases where structures become weak, enhanced, or inverted between views.

To separate radiometric and geometric difficulty, \textit{Synth} contains three displacement regimes. \textit{Synth-Extreme} applies object-wise non-rigid deformations using smooth grid-based flow fields, \textit{Synth-Perspective} simulates depth-dependent parallax under camera displacement, and \textit{Synth-Zero} contains no geometric motion and serves as a sanity test to evaluate whether spectral appearance changes alone cause a model to predict spurious displacement. Occlusions are generated by respecting object depth ordering during warping. Together, these regimes provide a controlled evaluation protocol for cross-spectral correspondence methods and complement real spectral acquisitions, where dense ground truth is typically unavailable.
To support reproducibility and benchmarking, we have released \textit{Synth} data and code at \url{https://github.com/fraunhoferhhi/XSpecCorr}

\subsection{Real Cross-Spectral Surgical and Biomedical Data} \label{sec:realdata}
In addition to quantitative benchmarks, we qualitatively assess cross-spectral correspondence on clinical data acquired with three complementary multi-sensor and/or multi-view setups:
(i) rectified or approximately rectified MSI stereo pairs \cite{tanriverdi2019dual,wisotzky2025real},
(ii) unrectified cross-spectral stereo pairs (e.g., RGB-SWIR and SWIR-SWIR) \cite{park2026multimodal},
and (iii) hyperspectral light-field sub-aperture view pairs \cite{maccormac2023lightfield,kray2025intraoperative}, where sub-images differ both in viewpoint and narrow-band filter response.
Across all regimes, the key challenge is that photometric consistency is violated by sensor- and wavelength-dependent radiometric responses, motivating correspondence learning that does not rely on brightness constancy.
These data capture severe modality gaps for which dense ground-truth correspondences are not available.
Therefore, the goal of this evaluation is to verify that the proposed cross-spectral correspondence framework yields visually stable and plausible displacement fields across challenging spectral combinations and viewpoints.

\textit{VIS to NIR MSI stereo. }
The first setup consists of a synchronized stereo pair of MSI snapshot mosaic cameras covering disjoint spectral ranges in the visible (VIS) and near-infrared (NIR) domain.
This yields cross-spectral stereo pairs with strong appearance changes between views due to non-overlapping wavelength sensitivity.
The VIS camera is sensitive between $[450,650]\,\mathrm{nm}$ capturing 16 spectral bands and the NIR camera is sensitive in the range of $[675,1000]\,\mathrm{nm}$ capturing 25 bands.
The image resolution is $1080{\times}2040$ pixels for each camera.
We construct evaluation pairs by selecting synchronized $I_L, I_R$ frames and projecting each multi-band capture to a single-channel intensity representation using the same normalization and intensity projection protocol as during training.

\textit{RGB to SWIR stereo (unrectified). }
The second setup is an unrectified stereo rig combining a conventional RGB camera with a SWIR camera operating at several distinct wavelength configurations in the $[1000,1700]\,\mathrm{nm}$ range using distinct illumination.
This enables two cross-spectral evaluation regimes: (i) RGB to SWIR$_k$ matching for each SWIR wavelength $k$, and (ii) SWIR$_i$ to SWIR$_j$ matching across different SWIR wavelengths.
The image resolution is $1080{\times}1440$ pixels for RGB and $1032{\times}1296$ pixels for SWIR. The images contain biochemical lab scenes captured with different working distances and/or different lens and illumination settings for each camera.
All images are converted to single-channel intensity inputs consistent with the cross-spectral backbone.
This setup specifically tests the ability to estimate a full 2D correspondence field under a strong modality gap and without relying on rectification.

\textit{HSI light-field data. }
This setup is a HSI light-field camera in which 66 sub-
images exhibit both viewpoint changes and narrow-band spectral filtering.
Each sub-aperture view corresponds to a different spectral filter with $\pm10\,\mathrm{nm}$ bandwidth between $[350,1000]\,\mathrm{nm}$. The resolution of each sub-view is $400{\times}400$ pixels. 
We form evaluation pairs from spectrally different sub-views (single-channel intensity), including both small- and larger-baseline view offsets, to probe correspondence robustness under simultaneous angular and spectral changes.

\subsection{Training and Evaluation Protocol}
\label{sec:method_training}

\begin{table}[!b]
\tiny
\caption{Training hyperparameters for our cross-spectral models.}
\centering
\begin{tabular}{r|ccc}
\toprule
  & \textit{Chairs} & \textit{Things} & \textit{Things, Sintel, KITTI, HD1K} \\
  & (Stage 1) & (Stage 2) & (Stage 3) \\
\midrule
Steps	      & 120k      & 150k      & 200k    \\
Learning Rate & 2.5e-4    & 1.5e-4    & 1.5e-4  \\
WDecay        & 1.0e-4    & 1.0e-5    & 1.0e-5  \\
Gamma         & 0.8       & 0.8       & 0.85    \\
Batch Size    & 8         & 8         & 8       \\
Size          & $368{\times}496$ & $400{\times}720$ & $368{\times}768$ \\
\bottomrule
\end{tabular}
\label{tab:OFhp}
\end{table}

In order to illustrate the practical efficacy of our framework, we train with standard supervised dense correspondence objectives on benchmark datasets providing ground-truth displacement. Specifically, we minimize a weighted sum of end-point error (EPE) across the iterative refinement sequence with exponential weighting as commonly used. 
For evaluation of the original baseline models, we use the official pretrained weights released by the respective authors. In contrast, all cross-spectral variants are trained from random initialization according to the following training protocol using identical data, augmentation, and optimization settings to enable fair comparisons.
We follow the standard multi-stage training practice used in literature for comparability.
We initially train on modified \textit{FlyingChairs} \cite{dosovitskiy2015flownet} and \textit{FlyingThings3D} \cite{mayer2016large}, followed on a mixture of modified \textit{MPI-Sintel} \cite{Butler:ECCV:2012}, \textit{HD1K} \cite{kondermann2016hci}, and \textit{KITTI} \cite{Menze2015CVPR}. All stages start from random initialization (no RGB-pretrained weights).
We keep hyperparameters aligned with the respective original training recipes (cf.~\cref{tab:OFhp}) and apply the same cross-spectral input protocol and augmentation to all models. 
Evaluation is performed quantitatively on the original and modified benchmarks as well as on our \textit{Synth} using the metrics defined below, and qualitatively on different real-world data.

We evaluate on the above mentioned modified benchmark sets as well as our synthetic cross-spectral dataset \textit{Synth} (Sec.~\ref{sec:method_synth}), where we additionally ablate the individual components of the cross-spectral training protocol.
Qualitatively, we run the trained correspondence networks and visualize the predicted displacement fields 
for each clinical setup. We inspect correspondence plausibility by checking structural alignment across object boundaries, the consistency of displacement discontinuities at depth edges, and the stability of predictions in low-texture regions where cross-spectral photometric cues are weak.
We follow standard validation protocols and use a fixed validation split in our setup with $10\,\%$ for \textit{KITTI}, no validation split for \textit{HD1K}, and $5\,\%$ for all other datasets.
We report EPE as the primary metric for dense correspondence, computed between predicted $\mathbf{u}$ and ground truth $\mathbf{u}'$.

\section{Results}
\begin{table}[t]
\tiny
\setlength{\tabcolsep}{4pt}
\caption{All models are evaluated using images from original benchmark datasets (RGB for the original and intensity for the cross-spectral models). The reported metric is EPE with column-wise (i.e., per dataset) marked \colorbox{yellow}{best} and \colorbox{gray!50}{second best} results.}
\centering
\begin{tabular}{lcrrrrrr}
\toprule
Architecture              & Variation     & Sintel& Sintel& KITTI & Middlebury & ETH3D & InStereo2K \\
                          &               & clean & final &       &       &       &        \\
\midrule
\multirow{2}{*}{RAFT}     & original      & 0.139 & 0.153 & 0.904 & 0.296 & 0.403 & 7.183  \\
                          & cross-spectral& 0.158 & 0.177 & 1.022 & 0.308 & 0.395 & 5.984  \\ \hline
\multirow{2}{*}{DIP}      & original      & \colorbox{gray!50}{0.121} & 0.161 & 1.762 & 0.260 & 0.381 & 5.941  \\
                          & cross-spectral& 0.132 & 0.162 & 0.889 & \colorbox{gray!50}{0.237} & \colorbox{yellow}{0.338} & 5.898  \\ \hline
\multirow{2}{*}{GMA}      & original      & 0.128 & 0.139 & 0.911 & 0.281 & 0.437 & 6.339  \\
                          & cross-spectral& 0.170 & 0.184 & 1.013 & 0.304 & 0.431 & 6.006  \\ \hline
\multirow{2}{*}{SEA-RAFT} & original      & \colorbox{yellow}{0.085} & \colorbox{yellow}{0.094} & \colorbox{yellow}{0.487} & \colorbox{yellow}{0.206} & 1.453 & 10.048 \\
                          & cross-spectral& 0.130 & 0.150 & 0.963 & 0.268 & \colorbox{gray!50}{0.371} & 12.300 \\ \hline
\multirow{2}{*}{SKFlow}   & original      & 0.122 & \colorbox{gray!50}{0.133} & \colorbox{gray!50}{0.782} & 0.258 & 0.409 & \colorbox{gray!50}{5.741}  \\
                          & cross-spectral& 0.157 & 0.179 & 1.052 & 0.293 & 0.410 & \colorbox{yellow}{5.714}  \\
\bottomrule
\end{tabular}
\label{tab:NetAna_orig}
\end{table}

The evaluation of the five cross-spectral network architectures is summarized in \cref{tab:NetAna_orig,tab:NetAna}. To assess whether the proposed framework preserves the baseline performance of the underlying architectures, all models trained under the cross-spectral protocol are additionally evaluated on the original benchmark datasets. The results, reported in \cref{tab:NetAna_orig}, show that the cross-spectral models maintain comparable performance to their respective baseline architectures.

Overall, cross-spectral models perform well on the available benchmark data\-sets. Due to the strong benchmark properties of \textit{Sintel} and \textit{KITTI}, all original architectures are optimized for these data. This is noticeable in the analyses using the original RGB datasets. For all cross-spectral models, the mean deviation from the respective originals for all datasets is $11.48\,\%$. The total mean deviation is $0.032$ EPE for \textit{MPI-Sintel}, $0.022$ EPE for \textit{Middlebury Stereo}, and $0.019$ EPE for \textit{KITTI}. For \textit{ETH3D} and \textit{InStereo2K}, the results are somewhat more diverse, with most of the cross-spectral models performing better than their original architectures resulting in total mean deviation of $-0.228$ EPE and $0.130$ EPE, respectively. 

\begin{table}[t]
\tiny
\setlength{\tabcolsep}{1.5pt}
\caption{The cross-spectral (cross-spec) and original models evaluated using modified evaluation images. The reported metric is EPE and \colorbox{yellow}{best}/\colorbox{gray!50}{second best} result marked.}
\centering
\begin{tabular}{ll|rrrrrrrrr}
\toprule
Architecture              & Variation     & Chairs           & Things           & Things           & Sintel          & Sintel          & KITTI            & Middle- & ETH3D& InStereo2K\\
                          &               &                  & clean            & final            & clean           & final           &                  & bury   &        &        \\
\midrule
\multirow{2}{*}{RAFT}     & original      & 15.619           & 28.581           & 30.268           & 2.473           & 2.870           & 17.797                                     & 14.947           & 7.193            & 29.145 \\
                          & cross-spec& 1.404            & 9.822            & 11.168           & 0.173           & 0.188           & 1.053            & 0.323            & 0.403            & 5.976  \\ \hline
\multirow{2}{*}{DIP}      & original      & 5.010            & 27.753           & 30.118           & 1.490           & 1.631           & 17.272                                     & 3.646            & 3.113            & 18.472 \\
                          & cross-spec& \colorbox{gray!50}{1.285}& 10.494           & 12.041           & \colorbox{yellow}{0.138}  &\colorbox{gray!50}{0.163}& \colorbox{yellow}{0.920}  & \colorbox{yellow}{0.243}   & \colorbox{yellow}{0.344}   & \colorbox{gray!50}{5.926}  \\ \hline
\multirow{2}{*}{GMA}      & original      & 15.636           & 36.477           & 35.813           & 3.342           & 2.874           & 15.480                                     & 7.578            & 7.323            & 31.563 \\
                          & cross-spec& 1.394            & 9.478            & 10.390           & 0.179           & 0.194           & 1.067            & 0.310            & 0.431            & 6.045  \\ \hline
\multirow{2}{*}{SEA-RAFT} & original      & 9.499            & 29.623           & 33.064           & 6.518           & 4.606           & 4.196                                      & 13.507           & 7.078            & 40.511 \\
                          & cross-spec& 1.331            & \colorbox{gray!50}{9.114}& \colorbox{gray!50}{9.334}&\colorbox{gray!50}{0.142}& \colorbox{yellow}{0.159}  & \colorbox{gray!50}{0.968}& \colorbox{gray!50}{0.280}& \colorbox{gray!50}{0.389}& 12.437 \\ \hline
\multirow{2}{*}{SKFlow}   & original      & 5.456            & 26.041           & 30.641           & 0.964           & 1.543           & 8.018                                      & 2.894            & 3.312            & 22.463 \\
                          & cross-spec& \colorbox{yellow}{1.242}   & \colorbox{yellow}{8.581}   & \colorbox{yellow}{9.002}   & 0.166           & 0.188           & 1.041           & 0.307            & 0.417            & \colorbox{yellow}{5.762} \\
\bottomrule
\end{tabular}
\label{tab:NetAna}
\end{table}

An architecture-specific analysis of all datasets reveals that the DIP-based model shows an average improvement of $10.12\,\%$ across all evaluated benchmark data as in four datasets the cross-spectral model performed better than the original. The remaining cross-spectral models show a slight degree of deterioration, with the most minimal deterioration observed in the RAFT-based model ($4.63\,\%$), followed by the GMA-based model with $12.99\,\%$, the SKFlow-based model with $18.52\,\%$, and the SEA-RAFT-based model with $31.3\,\%$. Thus, the cross-spectral models deliver results that are competitive to or, for some databases and/or architectures, better than the original architectures on conventional RGB inputs, albeit with architecture-dependent trade-offs.

The evaluation on the cross-spectral input data, i.e., the modified benchmark datasets, shows much better performances for the cross-spectral architectures compared to the originals, cf.~\cref{tab:NetAna}.
All cross-spectral models achieve the same accuracy rates on the cross-spectral datasets as on the original RGB benchmark data, while the original models are not able to achieve reasonable results on the cross-spectral data (about one order of magnitude worse).
The best models are the cross-spectral models from DIP, SKFlow, and SEA-RAFT architecture. The SEA-RAFT-based model is best in one and second in six datasets, while the SKFlow-based model is best in four datasets, and the DIP-based model is also best in four and second in three datasets. 

\begin{table}[t]
\tiny
\setlength{\tabcolsep}{4pt}
\caption{Model comparison on \textit{Synth} dataset regimes (preprocessed inputs). The \colorbox{yellow}{best} and \colorbox{gray!50}{second best} result for each displacement regime (column) is marked.}
\centering
\begin{tabular}{ll|rrr}
\toprule
Architecture              & Variation & Synth-Extreme & Synth-Perspective & Synth-Zero \\ 
\midrule
\multirow{2}{*}{RAFT}     & original      & 53.107  & 40.682            & 16.423     \\
                          & cross-spectral& 7.389   & 6.054             & 0.108      \\ \hline
\multirow{2}{*}{DIP}      & original      & 51.521  & 36.628            & 13.214     \\
                          & cross-spectral& 8.491   & 7.757             & \colorbox{yellow}{0.015} \\ \hline
\multirow{2}{*}{GMA}      & original      & 67.632  & 53.030            & 27.697 \\
                          & cross-spectral& 9.305   & 7.097             & 0.076 \\ \hline
\multirow{2}{*}{SEA-RAFT} & original      & 100.450 & 57.816            & 27.707 \\
                          & cross-spectral& \colorbox{gray!50}{6.653} & \colorbox{yellow}{4.389} & \colorbox{gray!50}{0.018} \\ \hline
\multirow{2}{*}{SKFlow}   & original      & 44.515  & 35.621            & 13.297 \\
                          & cross-spectral& \colorbox{yellow}{6.182} & \colorbox{gray!50}{4.620} & 0.164 \\
\bottomrule
\end{tabular}
\label{tab:NetAnaSynth}
\end{table}

\begin{table}[b]
\tiny
\setlength{\tabcolsep}{4pt}
\caption{Model comparison on \textit{Synth} using different spectral camera pairings. The values are averaged over all three \textit{Synth} displacement regimes. The reported metric is EPE ($\downarrow$), and the \colorbox{yellow}{best}/\colorbox{gray!50}{second best} result for each wavelength-pairing are marked.}
\centering
\begin{tabular}{ll|rrrrrrrrr}
\multirow{3}{*}{\parbox[c]{.97cm}{Architec-\\ture}} & \multirow{3}{*}{Variation} & LWIR & LWIR & MWIR & NIR & NIR & NIR & SWIR & SWIR & UV \\
 & & vs. & vs. & vs. & vs. & vs. & vs. & vs. & vs. & vs. \\
 & & MWIR & SWIR & SWIR & SWIR & UV & VIS & UV & VIS & VIS \\
\hline \hline
\multirow{2}{*}{RAFT}     & original  & 18.635 & 33.585 & 39.942 & 21.584 & 21.501 & 14.284 & 28.007 & 25.393 & 17.512 \\
                          & cross-spec& \colorbox{gray!50}{9.550}  & \colorbox{yellow}{9.889}  & \colorbox{yellow}{9.999}  & 6.581  & \colorbox{gray!50}{9.969}  & \colorbox{gray!50}{4.890}  & \colorbox{yellow}{9.795}  & 8.752  & \colorbox{gray!50}{8.511}  \\ \hline
\multirow{2}{*}{DIP}      & original  & 27.509 & 42.393 & 42.034 & 25.949 & 26.301 & 17.158 & 31.235 & 31.227 & 23.646 \\
                          & cross-spec& 12.821 & 13.469 & 13.331 & 9.716  & 12.311 & 7.277  & 12.457 & 11.063 & 11.574 \\ \hline
\multirow{2}{*}{GMA}      & original  & 21.687 & 41.808 & 43.557 & 24.476 & 27.261 & 16.840 & 32.712 & 31.186 & 21.123 \\
                          & cross-spec& 10.977 & 13.268 & 12.283 & 8.655  & 12.347 & 6.627  & 12.555 & 11.290 & 11.017 \\ \hline
SEA- & original  & 30.868 & 45.623 & 52.707 & 35.507 & 42.730 & 23.290 & 41.932 & 40.179 & 29.790 \\
RAFT                      & cross-spec& 10.697 & \colorbox{gray!50}{11.241} & 11.524 & \colorbox{yellow}{5.079}  & 12.658 & 5.591  & \colorbox{gray!50}{9.815}  & \colorbox{gray!50}{8.491}  & 8.721  \\ \hline
\multirow{2}{*}{SKFlow}   & original  & 20.550 & 41.417 & 43.370 & 18.666 & 22.217 & 12.715 & 27.138 & 25.091 & 16.766 \\
                          & cross-spec& \colorbox{yellow}{7.984}  & 11.302 & \colorbox{gray!50}{10.096} & \colorbox{gray!50}{6.524}  & \colorbox{yellow}{9.114}  & \colorbox{yellow}{4.779}  & 9.860  & \colorbox{yellow}{7.983}  & \colorbox{yellow}{7.726}  \\
\end{tabular}
\label{tab:RealSpectraSynth}
\end{table}

Finally, all models are evaluated on the proposed \textit{Synth} benchmark. 
The three displacement regimes in \cref{tab:NetAnaSynth} show that the cross-spectral training strategy consistently improves robustness under material-dependent spectral appearance changes. 
In the zero-motion regime, which serves as a sanity test for radiometrically induced false motion, both views differ only in their spectral appearance. The cross-spectral variants predict nearly zero displacement, whereas the original models introduce substantial spurious motion. 
For the other regimes, the cross-spectral models remain in a substantially lower error range than their original counterparts.

This trend is confirmed by the wavelength-pair analysis in \cref{tab:RealSpectraSynth}. 
Across all architectures and camera pairings, the cross-spectral variants achieve an average EPE of 9.78, while the original models show substantially higher EPE and stronger wavelength-dependent variation. 
The error of the original models increases with spectral separation, indicating sensitivity to material-dependent radiometric changes across distant wavelength ranges. 
In contrast, the cross-spectral variants remain stable, suggesting that the proposed augmentation strategy improves robustness across different correspondence backbones.

The ablation study in \cref{tab:ablation_searaft_realspectra} confirms that cross-spectral robustness results from the combination of sensor-agnostic input projection and spectral-response modulation. 
While the standard models perform poorly on real-spectra \textit{Synth}, a shared single-channel representation already reduces EPE. 
A larger improvement is obtained with view-dependent channel projections, or nonlinear radiometric transformations. 
The final configuration achieves the lowest EPE, indicating that view-dependent channel selection and nonlinear radiometric modulation provide complementary forms of spectral variability; the first introduces spatially heterogeneous, content-dependent contrast changes, whereas the last additionally exposes the model to strong nonlinear mappings and contrast inversions.

\begin{table}[t]
    \tiny
    \centering
    \caption{Ablation study of the cross-spectral training strategy for SEA-RAFT on the real-spectra \textit{Synth} data. Results are averaged over all evaluated spectral pairings. The reported metric is EPE ($\downarrow$) and \colorbox{yellow}{best}/\colorbox{gray!50}{second best} results are marked.}
    \label{tab:ablation_searaft_realspectra}
    \begin{tabular}{l c c c c c}
        Training & \multirow{2}{*}{Grayscale} & Channel & Monotonic & Monotonic & \multirow{2}{*}{EPE $\downarrow$} \\
        Configuration & & Mixture & Increasing & Decreasing & \\ \hline
        Standard RGB & & & & & 47.5810 \\
        grayscale-same & $\checkmark$ & & & & 32.0920 \\
        grayscale-different & $\checkmark$ & $\checkmark$ & & & \colorbox{gray!50}{16.0358} \\
        function-increase & $\checkmark$ & & $\checkmark$ & & 25.9549 \\
        function-full & $\checkmark$ & & $\checkmark$ & $\checkmark$ & 17.0292 \\
        Final cross-spectral & $\checkmark$ & $\checkmark$ & $\checkmark$ & $\checkmark$ & \colorbox{yellow}{9.3132} \\
    \end{tabular}
\end{table}

\subsection{Analysis of the Image-specific Features}
To verify that the performance gains on cross-spectral inputs are reflected at the representation level, we analyze the image-specific features produced by the correspondence backbones. Specifically, we study whether features extracted from the two views $(I_L,I_R)$ remain consistent under spectral mismatch, as this directly affects the sharpness and uniqueness of the subsequent matching signal.

\begin{table}[b]
\tiny
\centering
\caption{Cosine feature similarity across data processing variants and real-spectral \textit{Synth} (last column) for different architectures (original and cross-spectral). The data modification by non-overlapping channel pairing is labeled as \emph{Permute}, while the spectral response modulation using the function family is referred to as \emph{Monotonic Increase} (no inversion), \emph{Monotonic Decrease} (inversion), \emph{Mixed monotonic} (random inversion).}
\label{tab:processing_adaptation}
\begin{tabular}{ll|ccccc|c}
\toprule
\multicolumn{2}{r|}{\multirow{2}{*}{Input Variation}} & \multirow{2}{*}{Identity} & Permute  & Monotonic  & Monotonic  & Mixed & Real-Spectral\\ 
                          &           &        & Channels & Increasing & Decreasing & Monotonic & \textit{Synth} \\ 
\midrule
\multirow{2}{*}{SEA-RAFT} & original      & 1      & 0.9277 & 0.7813 & 0.7822 & 0.2540 & 0.3999\\
                          & cross-spectral& 1      & 0.9864 & 0.9463 & 0.9384 & 0.9359 & 0.7537\\ \hline
\multirow{2}{*}{RAFT}     & original      & 1      & 0.9553 & 0.8969 & 0.8793 & 0.1005 & 0.3581\\
                          & cross-spectral& 1      & 0.9930 & 0.9279 & 0.9278 & 0.9223 & 0.5517\\ \hline
\multirow{2}{*}{SKFlow}   & original      & 1      & 0.9443 & 0.8740 & 0.8723 & 0.1274 & 0.3688\\
                          & cross-spectral& 1      & 0.9937 & 0.9422 & 0.9384 & 0.9220 & 0.5613\\ \hline
\multirow{2}{*}{GMA}      & original      & 1      & 0.9490 & 0.8829 & 0.8867 & 0.0970 & 0.3645\\
                          & cross-spectral& 1      & 0.9948 & 0.9434 & 0.9301 & 0.9164 & 0.5457\\ \hline
\multirow{2}{*}{DIP}      & original      & 1      & 0.9237 & 0.8443 & 0.8395 & 0.2867 & 0.4395\\
                          & cross-spectral& 1      & 0.9753 & 0.9120 & 0.9137 & 0.9027 & 0.6093\\
\bottomrule
\end{tabular}
\end{table}

For a randomly sampled subset of benchmark images used for evaluation, we extract the backbone feature maps 
and compute their cosine similarity, averaged over spatial locations, as a diagnostic measure of radiometric invariance. It is important to note, that we use the same viewpoint for both inputs $(I_L,I_R)$ to return similarity differences to its cross-spectral behavior only and is not expected to rank correspondence accuracy across architectures. \cref{tab:processing_adaptation} reports the resulting cosine feature similarity for different input variants (identity, non-overlapping channel pairing, and spectral-response modulation) of these images.

Across all five backbones, the original architectures exhibit a pronounced sensitivity to radiometric inversions. While channel permutation and monotonic intensity remappings using our introduced function family without and with a fixed inversion retain moderate-to-high cosine feature similarity ($\bar{r}=0.94$ for \emph{Permute}, $\bar{r}=0.85$ for \emph{Monotonic Increasing/Decreasing}), the \emph{Mixed} setting collapses the similarity to $\bar{r}=0.17$ (range $[0.10,0.29]$). In contrast, the cross-spectral architectures maintain consistently high feature similarity across all variants ($r>0.9$). The strongest improvement is observed for \emph{Mixed Monotonic}, where our cross-spectral training increases similarity by $0.75$ on average, indicating substantially improved invariance to modality-induced radiometric shifts.

The same trend is observed in our \textit{Synth} dataset. \cref{tab:processing_adaptation}(last column) shows that cross-spectral training increases cosine feature similarity for all architectures from a mean of $\bar{r}=0.386$ (original) to $\bar{r}=0.604$ (cross-spectral). The largest improvement and highest similarity is observed for cross-spectral SEA-RAFT. The cross-spectral DIP achieves the second-highest cosine feature similarity, although its improvement relative to the baseline (original RGB trained models) is the smallest among the evaluated architectures. These trends are consistent with the performance gains observed on \textit{Synth} (\cref{tab:NetAnaSynth}), particularly for \textit{Synth-Zero}, where no geometric displacement is present and any predicted motion therefore directly reflects sensitivity to spectral/radiometric mismatch.
This indicates that enforcing a sensor-agnostic interface and explicitly training for radiometric invariance addresses a central bottleneck for cross-spectral acquisition systems. The original models retain high feature similarity for mild spectral perturbations, but correlations collapse at mixed radiometric changes, directly increasing ambiguity in subsequent matching signal. In contrast, cross-spectral models maintain consistently high feature similarity for all variants. 

\begin{figure}[tb]
  \tiny
  \centering
  \begin{tabular}{cccccc}
    \rotatebox[origin=ll]{90}{MSI} & \rotatebox[origin=ll]{90}{stereo} &
    \includegraphics[width=0.23\columnwidth]{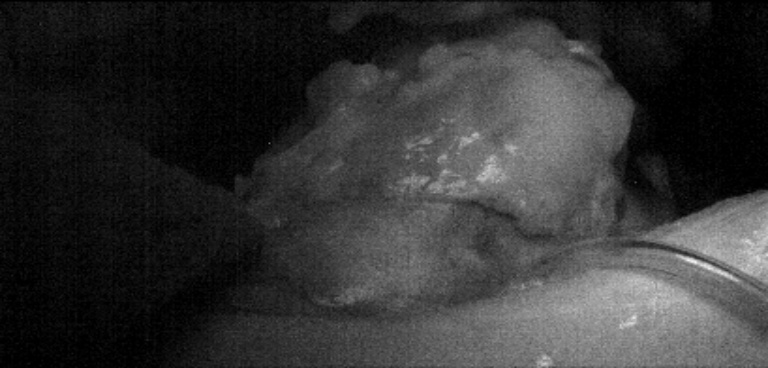} &
    \includegraphics[width=0.23\columnwidth]{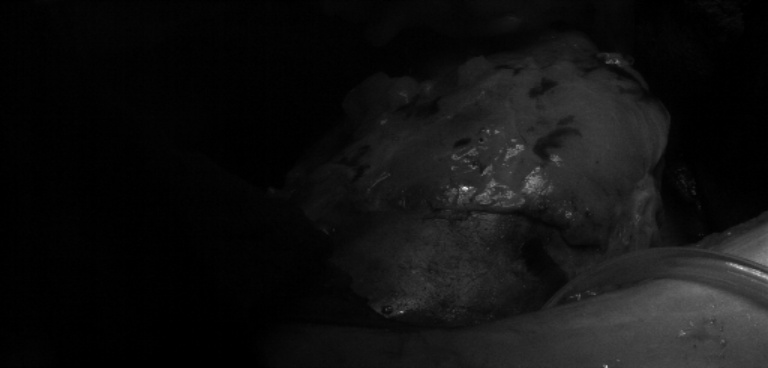} &
    \includegraphics[width=0.23\columnwidth]{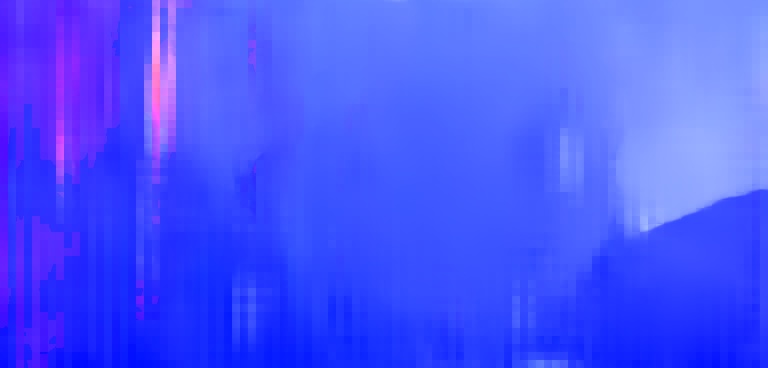} &
    \includegraphics[width=0.23\columnwidth]{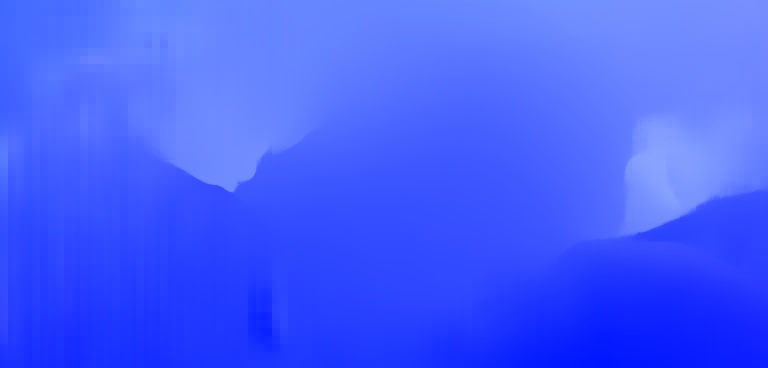} \\
    
    \rotatebox[origin=ll]{90}{RGB-} & \rotatebox[origin=ll]{90}{SWIR} &
    \includegraphics[width=0.23\columnwidth, height=0.1\columnwidth]{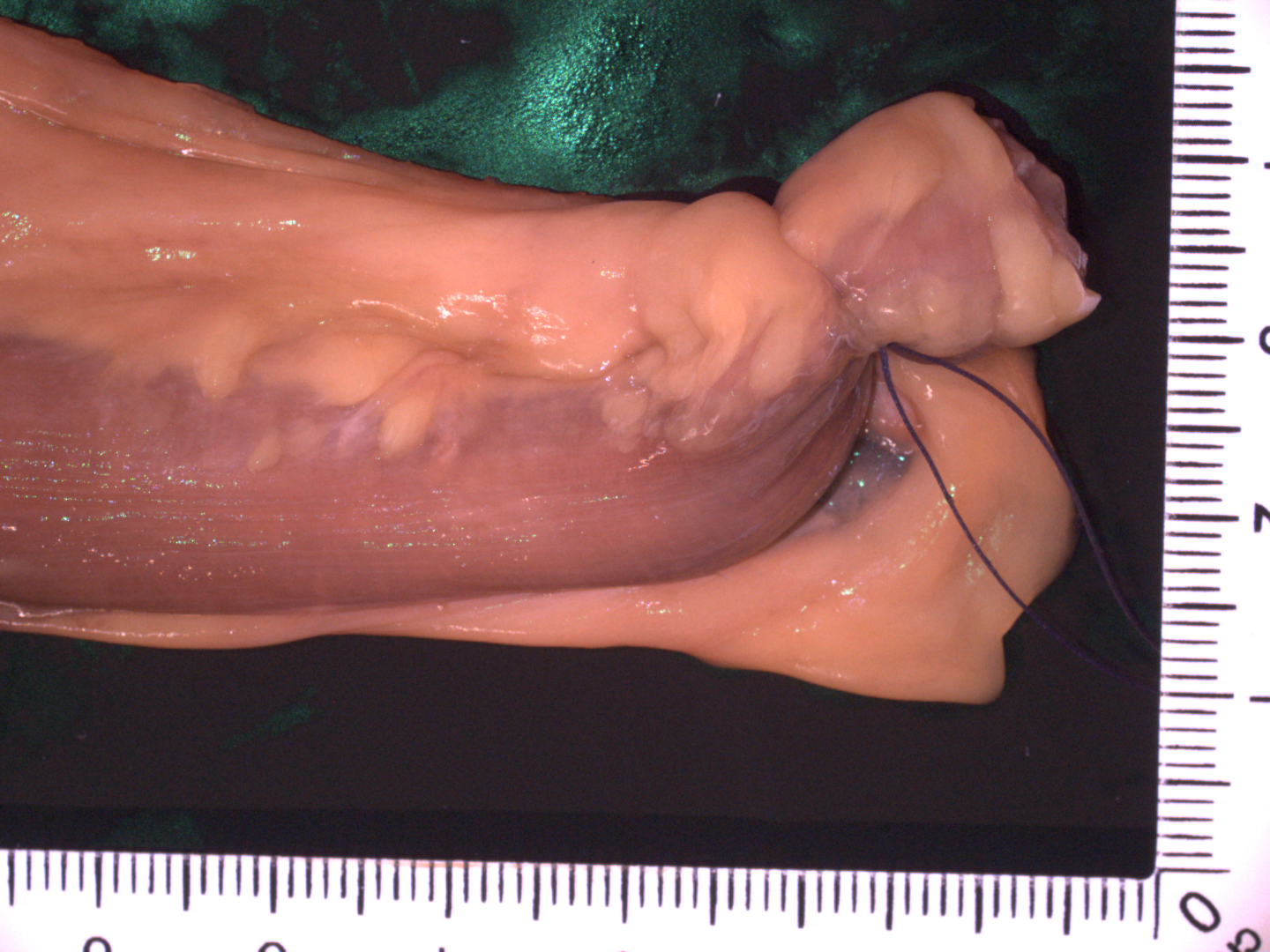} &
    \includegraphics[width=0.23\columnwidth, height=0.1\columnwidth]{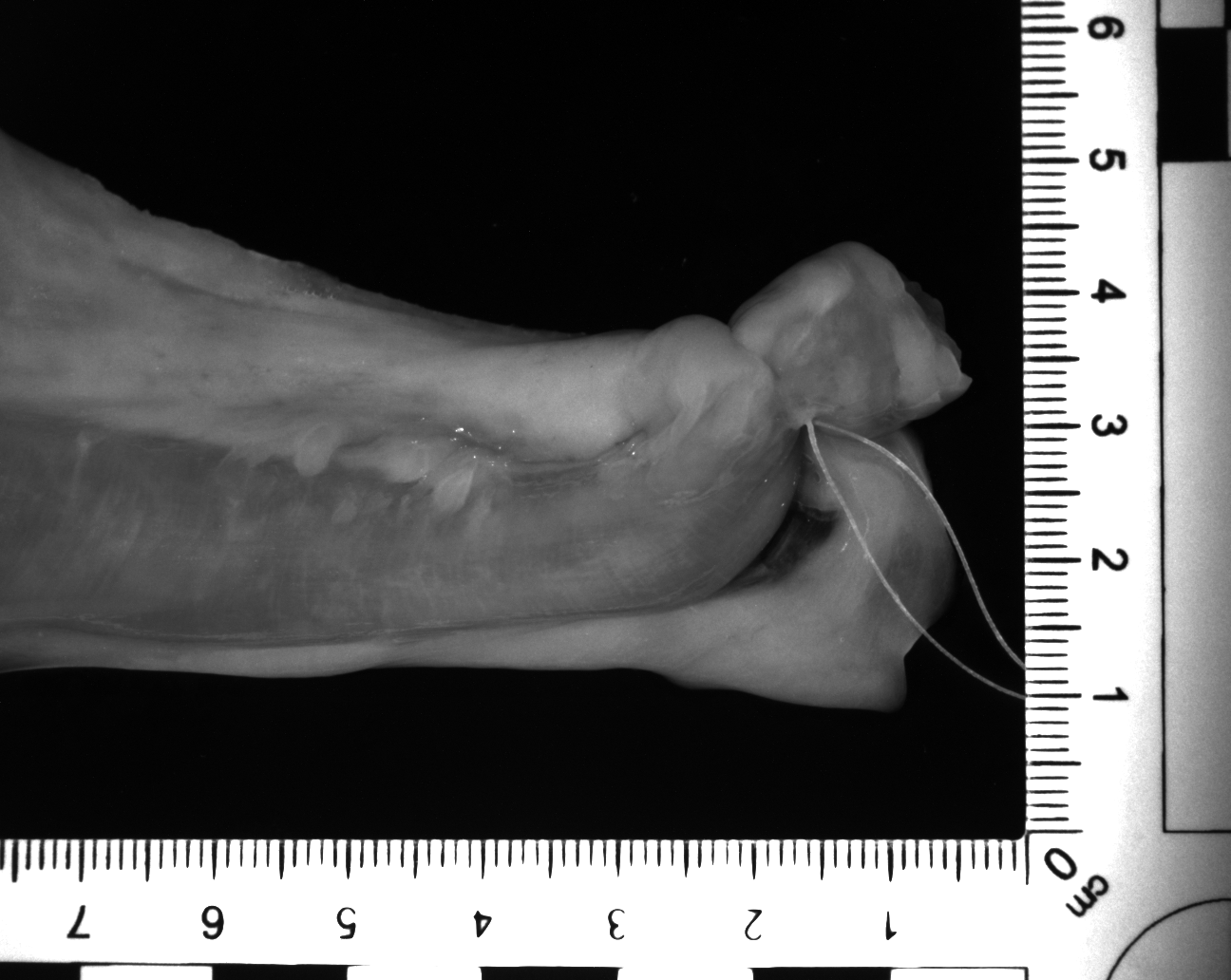} &
    \includegraphics[width=0.23\columnwidth]{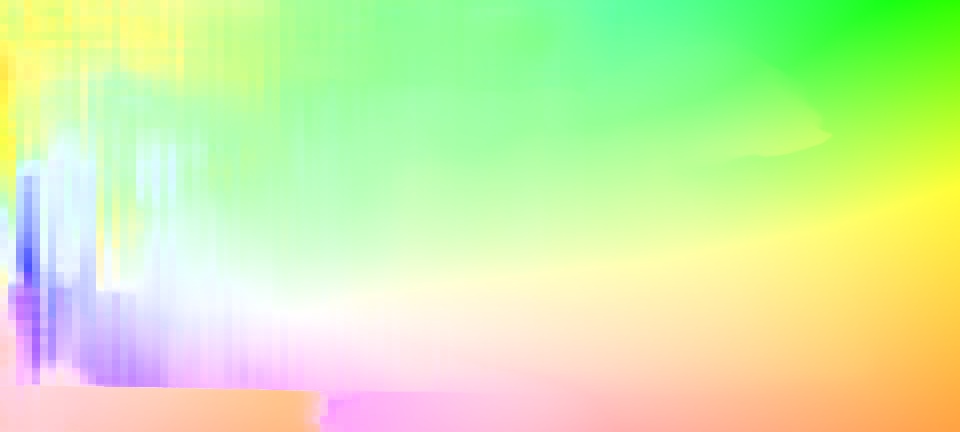} &
    \includegraphics[width=0.23\columnwidth]{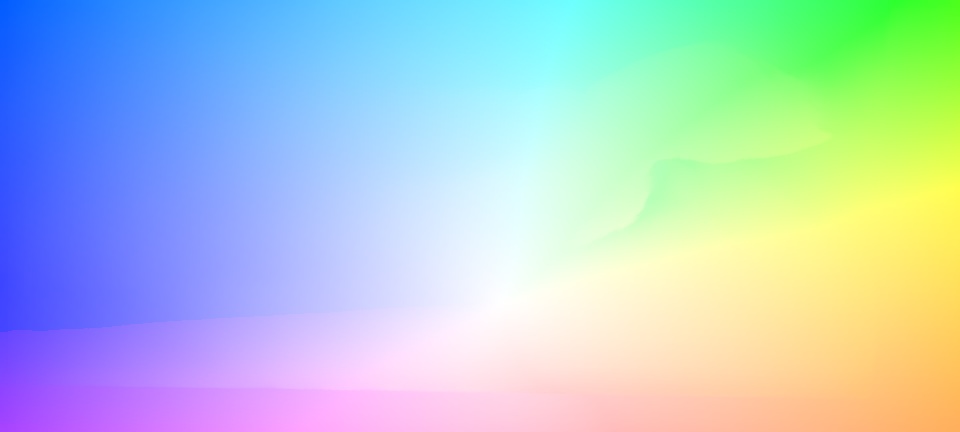} \\
    
    \rotatebox[origin=ll]{90}{HSI} & \rotatebox{90}{light-field} &
    \includegraphics[width=0.23\columnwidth]{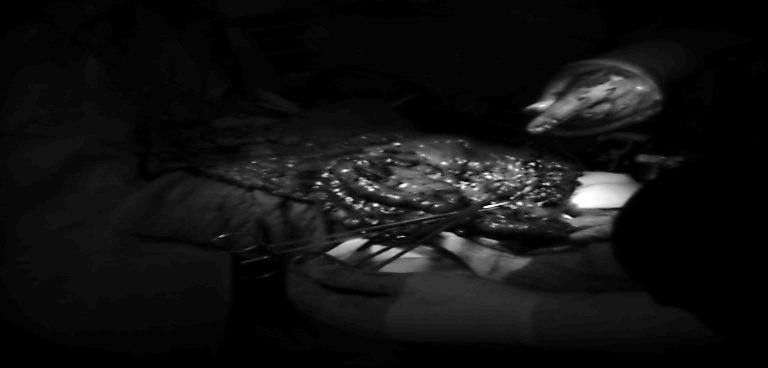} &
    \includegraphics[width=0.23\columnwidth]{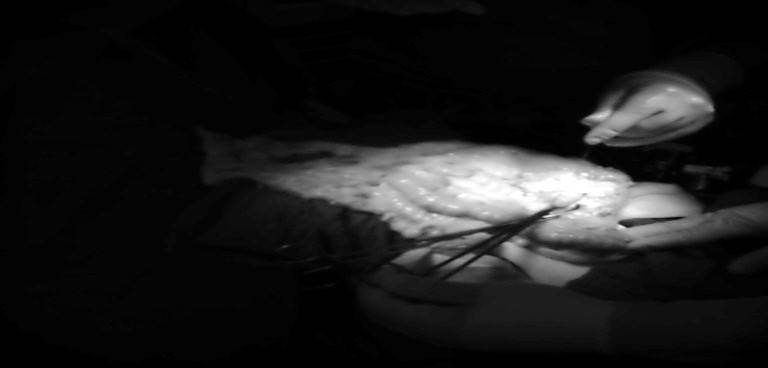} & 
    \includegraphics[width=0.23\columnwidth]{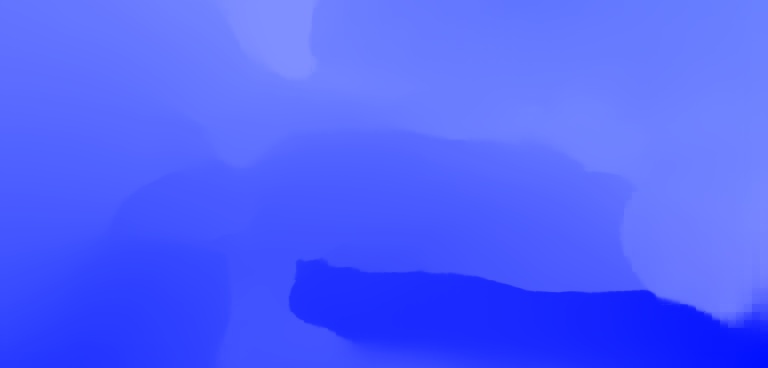} &
    \includegraphics[width=0.23\columnwidth]{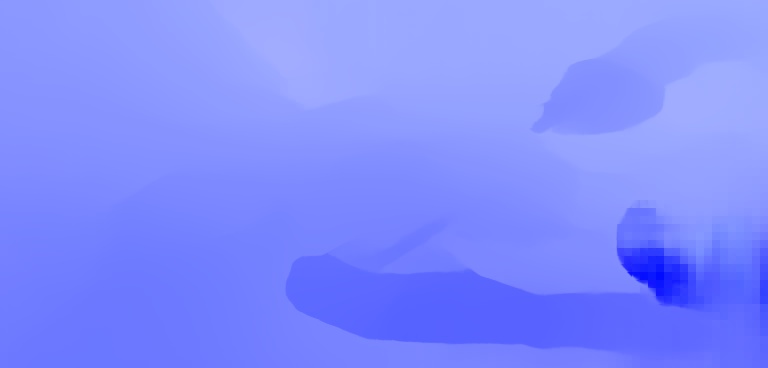} \\
    
    && \multirow{2}{*}{View $I_a$} & \multirow{2}{*}{View $I_b$} & Displacement Map & Displacement Map \\
    &&           &            & Original         & Cross-Spectral \\
  \end{tabular}
  \caption{Different real-world data acquired with three different cross-spectral setups. All displacement maps are generated with the SEA-RAFT-based architecture (original and cross-spectral). The results of the other architectures can be found in the Supplementary Material. 
  Top row: Surgical images acquired with an VIS-NIR MSI-stereo setup. Middle row: Tissue acquired with unrectified RGB-SWIR setup. Bottom row: Two surgical sub-views acquired with a HSI light-field setup. All displacement fields from the original SEA-RAFT architecture contain strong artifact or miss content, while the cross-spectral variant produces more spatially coherent fields with fewer obvious artifacts.}
  \label{fig:realdata}
\end{figure}

Although no ground-truth correspondences are available for the used real-world data, the qualitative results are consistent with the quantitative trends. The cross-spectral models produce coherent displacement fields across all different heterogeneous setups, while the original models are more prone to modality-specific artifacts under strong spectral mismatch, see \cref{fig:realdata}. Together with the representation analysis, this supports the practical utility of the proposed training protocol as an enabling component for multi-sensor alignment.

\section{Conclusion}
We presented a data-centric framework for cross-spectral correspondence benchmark generation and training. Instead of designing a task-specific architecture, the proposed approach focuses on constructing realistic cross-spectral training and evaluation data through sensor-agnostic spectral-response modulation. The resulting framework generates both cross-spectral variants of established benchmarks and the synthetic \textit{Synth} benchmark with dense correspondence ground truth under controlled spectral appearance variation, providing a systematic basis for training and evaluating models under spectral mismatch.

Across several modern correspondence architectures, the proposed training strategy substantially improves robustness under cross-spectral inputs while preserving competitive performance on RGB benchmarks. The results on \textit{Synth} indicate that conventional models do not primarily fail due to insufficient geometric matching capacity. Rather, their learned representations remain too dependent on appearance statistics observed during conventional RGB training. Ablation experiments show compatibility between training distribution and target spectral domain is a dominant factor for correspondence under spectral mismatch.

Beyond the dense correspondence task considered in this work, the proposed data generation framework provides a practical methodology for systematically training and benchmarking vision models under realistic cross-spectral conditions. Robust correspondence is an enabling component for multimodal spectral imaging applications, including spatially coherent spectral fusion, depth recovery, and tissue or perfusion analysis in image-guided surgery.
Since dense ground truth is rarely available for real surgical spectral data, generating realistic synthetic data and constructing benchmarks are effective ways to develop, compare, and validate correspondence models prior to deployment on clinical data.
Overall, our results demonstrate that realistic cross-spectral data generation is a key enabler for robust correspondence learning. We believe that the proposed framework is architecture-agnostic and provides a practical foundation for future research on cross-spectral vision, synthetic data generation, and multimodal medical imaging beyond specific architectures studied in this work.

\section*{Acknowledgements}
This work was funded by the BMFTR - Federal Ministry of Research, Technology and Space under the project REFRAME, grant no. 01IS2407A. All experiments involving humans were performed in accordance with the Declaration of Helsinki, and the protocols were approved by the Ethics committee of Charité - University Hospital Berlin.

%
%
\bibliographystyle{splncs04}
\bibliography{main}

@String(CVPR  = {IEEE Conf. Comput. Vis. Pattern Recog.})

@String(ECCV  = {Eur. Conf. Comput. Vis.})

@String(CVPR  = {CVPR})

@String(ECCV  = {ECCV})

@inproceedings{edstedt2024roma,
  title={Roma: Robust dense feature matching},
  author={Edstedt, Johan and Sun, Qiyu and B{\"o}kman, Georg and Wadenb{\"a}ck, M{\aa}rten and Felsberg, Michael},
  booktitle={2024 IEEE/CVF Conference on Computer Vision and Pattern Recognition (CVPR)},
  pages={19790--19800},
  year={2024},
  organization={IEEE}
}

@inproceedings{wang2024dust3r,
  title={Dust3r: Geometric 3d vision made easy},
  author={Wang, Shuzhe and Leroy, Vincent and Cabon, Yohann and Chidlovskii, Boris and Revaud, Jerome},
  booktitle={2024 IEEE/CVF Conference on Computer Vision and Pattern Recognition (CVPR)},
  pages={20697--20709},
  year={2024},
  organization={IEEE}
}

@article{lu2014medical,
  title={Medical hyperspectral imaging: a review},
  author={Lu, Guolan and Fei, Baowei},
  journal={Journal of biomedical optics},
  volume={19},
  number={1},
  pages={010901},
  year={2014},
  publisher={SPIE}
}

@article{seidlitz2022robust,
  title={Robust deep learning-based semantic organ segmentation in hyperspectral images},
  author={Seidlitz, Silvia and Sellner, Jan and Odenthal, Jan and {\"O}zdemir, Berkin and Studier-Fischer, Alexander and Kn{\"o}dler, Samuel and Ayala, Leonardo and Adler, Tim J and Kenngott, Hannes G and Tizabi, Minu and others},
  journal={Medical Image Analysis},
  volume={80},
  pages={102488},
  year={2022},
  publisher={Elsevier}
}

@article{muhle2021comparison,
  title={Comparison of different spectral cameras for image-guided organ transplantation},
  author={M{\"u}hle, Richard and Markgraf, Wenke and Hilsmann, Anna and Malberg, Hagen and Eisert, Peter and Wisotzky, Eric L},
  journal={Journal of biomedical optics},
  volume={26},
  number={7},
  pages={076007},
  year={2021},
  publisher={SPIE}
}

@article{wisotzky2025telepresence,
  title={Telepresence for surgical assistance and training using eXtended reality during and after pandemic periods},
  author={Wisotzky, Eric L and Rosenthal, Jean-Claude and Meij, Senna and van den Dobblesteen, John and Arens, Philipp and Hilsmann, Anna and Eisert, Peter and Uecker, Florian Cornelius and Schneider, Armin},
  journal={Journal of telemedicine and telecare},
  volume={31},
  number={1},
  pages={14--28},
  year={2025},
  publisher={SAGE Publications Sage UK: London, England}
}

@article{wisotzky2020validation,
  title={Validation of two techniques for intraoperative hyperspectral human tissue determination},
  author={Wisotzky, Eric L and Kossack, Benjamin and Uecker, Florian C and Arens, Philipp and Hilsmann, Anna and Eisert, Peter},
  journal={Journal of Medical Imaging},
  volume={7},
  number={6},
  pages={065001--065001},
  year={2020},
  publisher={Society of Photo-Optical Instrumentation Engineers}
}

@inproceedings{wisotzky2019validation,
  title={Validation of two techniques for intraoperative hyperspectral human tissue determination},
  author={Wisotzky, Eric L and Kossack, Benjamin and Uecker, Florian C and Arens, Philipp and Dommerich, Steffen and Hilsmann, Anna and Eisert, Peter},
  booktitle={Medical Imaging 2019: Image-Guided Procedures, Robotic Interventions, and Modeling},
  volume={10951},
  pages={511--525},
  year={2019},
  organization={SPIE}
}

@inproceedings{wisotzky2024MSIfusion,
  title={Multispectral Stereo-Image Fusion for 3D Hyperspectral Scene Reconstruction},
  author={Wisotzky, Eric L and Triller, Jost and Hilsmann, Anna and Eisert, Peter},
  booktitle={Proceedings of the 19th International Joint Conference on Computer Vision, Imaging and Computer Graphics Theory and Applications - Volume 3: VISAPP},
  volume={19},
  pages={88--99},
  year={2024},
  organization={SciTePress}
}

@article{wisotzky2018intraoperative,
  title={Intraoperative hyperspectral determination of human tissue properties},
  author={Wisotzky, Eric Larry and Uecker, Florian Cornelius and Arens, Philipp and Dommerich, Steffen and Hilsmann, Anna and Eisert, Peter},
  journal={Journal of biomedical optics},
  volume={23},
  number={9},
  pages={091409--091409},
  year={2018},
  publisher={Society of Photo-Optical Instrumentation Engineers}
}

@inproceedings{wisotzky2025continuous,
  title={Continuous hyperspectral stereo-imaging for image-guided surgery},
  author={Wisotzky, Eric L and Triller, Jost and Hilsmann, Anna and Eisert, Peter},
  booktitle={Advanced Biomedical and Clinical Diagnostic and Surgical Guidance Systems XXIII},
  volume={13306},
  pages={36--42},
  year={2025},
  organization={SPIE}
}

@article{wisotzky20233d,
  title={3D Hyperspectral Light-Field Imaging: a first intraoperative implementation},
  author={Wisotzky, Eric L and Hilsmann, Anna and Eisert, Peter},
  journal={Current Directions in Biomedical Engineering},
  volume={9},
  number={1},
  pages={611--614},
  year={2023}
}

@article{wisotzky2024automatic,
  title={Automatic tissue differentiation in parotidectomy using hyperspectral imaging},
  author={Wisotzky, Eric L and Schill, Alexander and Hilsmann, Anna and Eisert, Peter and Knoke, Michael},
  journal={Current Directions in Biomedical Engineering},
  volume={10},
  number={4},
  pages={682--685},
  year={2024}
}

@article{clancy2021intraoperative,
  title={Intraoperative colon perfusion assessment using multispectral imaging},
  author={Clancy, Neil T and Soares, Ant{\'o}nio S and Bano, Sophia and Lovat, Laurence B and Chand, Manish and Stoyanov, Danail},
  journal={Biomedical optics express},
  volume={12},
  number={12},
  pages={7556--7567},
  year={2021},
  publisher={Optical Society of America}
}

@inproceedings{sellner2023semantic,
  title={Semantic segmentation of surgical hyperspectral images under geometric domain shifts},
  author={Sellner, Jan and Seidlitz, Silvia and Studier-Fischer, Alexander and Motta, Alessandro and {\"O}zdemir, Berkin and M{\"u}ller-Stich, Beat Peter and Nickel, Felix and Maier-Hein, Lena},
  booktitle={International Conference on Medical Image Computing and Computer-Assisted Intervention},
  pages={618--627},
  year={2023},
  organization={Springer}
}

@article{azagra2023endomapper,
  title={Endomapper dataset of complete calibrated endoscopy procedures},
  author={Azagra, Pablo and Sostres, Carlos and Ferr{\'a}ndez, {\'A}ngel and Riazuelo, Luis and Tomasini, Clara and Barbed, O Le{\'o}n and Morlana, Javier and Recasens, David and Batlle, Victor M and G{\'o}mez-Rodr{\'\i}guez, Juan J and others},
  journal={Scientific Data},
  volume={10},
  number={1},
  pages={671},
  year={2023},
  publisher={Nature Publishing Group UK London}
}

@techreport{kokaly2017usgs,
  title={USGS spectral library version 7},
  author={Kokaly, Raymond F and Clark, Roger N and Swayze, Gregg A and Livo, K Eric and Hoefen, Todd M and Pearson, Neil C and Wise, Richard A and Benzel, William and Lowers, Heather A and Driscoll, Rhonda L and others},
  year={2017},
  institution={US Geological Survey}
}

@inproceedings{lipson2021raft-stereo,
  title={Raft-stereo: Multilevel recurrent field transforms for stereo matching},
  author={Lipson, Lahav and Teed, Zachary and Deng, Jia},
  booktitle={2021 International Conference on 3D Vision (3DV)},
  pages={218--227},
  year={2021},
  organization={IEEE}
}

@inproceedings{teed2020raft,
  title={Raft: Recurrent all-pairs field transforms for optical flow},
  author={Teed, Zachary and Deng, Jia},
  booktitle={European conference on computer vision},
  pages={402--419},
  year={2020},
  organization={Springer}
}

@article{xu2023unifying,
  title={Unifying flow, stereo and depth estimation},
  author={Xu, Haofei and Zhang, Jing and Cai, Jianfei and Rezatofighi, Hamid and Yu, Fisher and Tao, Dacheng and Geiger, Andreas},
  journal={IEEE Transactions on Pattern Analysis and Machine Intelligence},
  volume={45},
  number={11},
  pages={13941--13958},
  year={2023},
  publisher={IEEE}
}

@inproceedings{liu2020flow2stereo,
  title={Flow2stereo: Effective self-supervised learning of optical flow and stereo matching},
  author={Liu, Pengpeng and King, Irwin and Lyu, Michael R and Xu, Jia},
  booktitle={Proceedings of the IEEE/CVF conference on computer vision and pattern recognition},
  pages={6648--6657},
  year={2020}
}

@article{wisotzky2025real,
  title={Real-time fusion of stereo vision and hyperspectral imaging for objective decision support during surgery},
  author={Wisotzky, Eric L and Triller, Jost and Knoke, Michael and Globke, Brigitta and Hilsmann, Anna and Eisert, Peter},
  journal={Computer Vision and Image Understanding},
  pages={104541},
  year={2025},
  publisher={Elsevier}
}

@article{park2026multimodal,
  title={Multimodal Shortwave Infrared Imaging for Visualization of Injection Laryngoplasty},
  author={Park, Roy K and Lee, Melissa C and H{\"a}rtl, Simon and Ar{\'u}s, Bernardo A and Nuyen, Brian and Sung, Chih-Kwang and Baik, Fred M and Bruns, Oliver T and Valdez, Tulio A},
  journal={Otolaryngology--Head and Neck Surgery},
  volume={174},
  number={1},
  pages={185--194},
  year={2026},
  publisher={Wiley Online Library}
}

@article{maccormac2023lightfield,
  title={Lightfield hyperspectral imaging in neuro-oncology surgery: an IDEAL 0 and 1 study},
  author={MacCormac, Oscar and Noonan, Philip and Janatka, Mirek and Horgan, Conor C and Bahl, Anisha and Qiu, Jianrong and Elliot, Matthew and Trotouin, Th{\'e}o and Jacobs, Jaco and Patel, Sabina and others},
  journal={Frontiers in Neuroscience},
  volume={17},
  pages={1239764},
  year={2023},
  publisher={Frontiers Media SA}
}

@inproceedings{zheng2022dip,
  title={Dip: Deep inverse patchmatch for high-resolution optical flow},
  author={Zheng, Zihua and Nie, Ni and Ling, Zhi and Xiong, Pengfei and Liu, Jiangyu and Wang, Hao and Li, Jiankun},
  booktitle={Proceedings of the IEEE/CVF Conference on Computer Vision and Pattern Recognition},
  pages={8925--8934},
  year={2022}
}

@article{sun2022skflow,
  title={Skflow: Learning optical flow with super kernels},
  author={Sun, Shangkun and Chen, Yuanqi and Zhu, Yu and Guo, Guodong and Li, Ge},
  journal={Advances in Neural Information Processing Systems},
  volume={35},
  pages={11313--11326},
  year={2022}
}

@inproceedings{wang2024sea,
  title={Sea-raft: Simple, efficient, accurate raft for optical flow},
  author={Wang, Yihan and Lipson, Lahav and Deng, Jia},
  booktitle={European Conference on Computer Vision},
  pages={36--54},
  year={2024},
  organization={Springer}
}

@inproceedings{brucker2024cross,
  title={Cross-spectral gated-rgb stereo depth estimation},
  author={Brucker, Samuel and Walz, Stefanie and Bijelic, Mario and Heide, Felix},
  booktitle={Proceedings of the IEEE/CVF Conference on Computer Vision and Pattern Recognition},
  pages={21654--21665},
  year={2024}
}

@inproceedings{zhi2018deep,
  title={Deep material-aware cross-spectral stereo matching},
  author={Zhi, Tiancheng and Pires, Bernardo R and Hebert, Martial and Narasimhan, Srinivasa G},
  booktitle={Proceedings of the IEEE conference on computer vision and pattern recognition},
  pages={1916--1925},
  year={2018}
}

@inproceedings{tuzcuouglu2024xoftr,
  title={Xoftr: Cross-modal feature matching transformer},
  author={Tuzcuo{\u{g}}lu, {\"O}nder and K{\"o}ksal, Aybora and Sofu, Bu{\u{g}}ra and Kalkan, Sinan and Alatan, A Aydin},
  booktitle={Proceedings of the IEEE/CVF conference on computer vision and pattern recognition},
  pages={4275--4286},
  year={2024}
}

@inproceedings{mehltretter2018multimodal,
  title={Multimodal dense stereo matching},
  author={Mehltretter, Max and Kleinschmidt, Sebastian P and Wagner, Bernardo and Heipke, Christian},
  booktitle={German Conference on Pattern Recognition},
  pages={407--421},
  year={2018},
  organization={Springer}
}

@inproceedings{tanriverdi2019dual,
  title={Dual snapshot hyperspectral imaging system for 41-band spectral analysis and stereo reconstruction},
  author={Tanriverdi, Fatih and Schuldt, Dennis and Thiem, J{\"o}rg},
  booktitle={International Symposium on Visual Computing},
  pages={3--13},
  year={2019},
  organization={Springer}
}

@inproceedings{kray2025intraoperative,
  title={Intraoperative perfusion assessment by continuous, low-latency hyperspectral light-field imaging: development, methodology, and clinical application},
  author={Kray, Stefan and Schmid, Andreas and Wisotzky, Eric L and Gerlich, Moritz and Apweiler, Sebastian and Hilsmann, Anna and Greiner, Thomas and Eisert, Peter and Kneist, Werner},
  booktitle={Advanced Biomedical and Clinical Diagnostic and Surgical Guidance Systems XXIII},
  volume={13306},
  pages={30--35},
  year={2025},
  organization={SPIE}
}

@article{genser2020camera,
  title={Camera array for multi-spectral imaging},
  author={Genser, Nils and Seiler, J{\"u}rgen and Kaup, Andr{\'e}},
  journal={IEEE Transactions on Image Processing},
  volume={29},
  pages={9234--9249},
  year={2020},
  publisher={IEEE}
}

@article{huang2024data,
  title={Data rectification and decoding of a microlens array-based multi-spectral light field imaging system},
  author={Huang, Yizhi and Hossain, Md Moinul and Liu, Yan and Sun, Kai and Zhang, Biao and Xu, Chuanlong},
  journal={Optics and Lasers in Engineering},
  volume={180},
  pages={108327},
  year={2024},
  publisher={Elsevier}
}

@inproceedings{Butler:ECCV:2012,
title = {A naturalistic open source movie for optical flow evaluation},
author = {Butler, D. J. and Wulff, J. and Stanley, G. B. and Black, M. J.},
booktitle = {European Conf. on Computer Vision (ECCV)},
editor = {{A. Fitzgibbon et al. (Eds.)}},
publisher = {Springer-Verlag},
series = {Part IV, LNCS 7577},
month = oct,
pages = {611--625},
year = {2012}
}

@inproceedings{Menze2015CVPR,
  title={Object scene flow for autonomous vehicles},
  author={Menze, Moritz and Geiger, Andreas},
  booktitle={IEEE conference on computer vision and pattern recognition},
  pages={3061--3070},
  year={2015}
}

@inproceedings{kondermann2016hci,
  title={The HCI Benchmark Suite: Stereo and Flow Ground Truth With Uncertainties for Urban Autonomous Driving},
  author={Kondermann, Daniel and Nair, Rahul and Honauer, Katrin and Krispin, Karsten and Andrulis, Jonas and Brock, Alexander and Gussefeld, Burkhard and Rahimimoghaddam, Mohsen and Hofmann, Sabine and Brenner, Claus and others},
  booktitle={IEEE Conference on Computer Vision and Pattern Recognition Workshops},
  pages={19--28},
  year={2016}
}

@inproceedings{mayer2016large,
  title={A large dataset to train convolutional networks for disparity, optical flow, and scene flow estimation},
  author={Mayer, Nikolaus and Ilg, Eddy and Hausser, Philip and Fischer, Philipp and Cremers, Daniel and Dosovitskiy, Alexey and Brox, Thomas},
  booktitle={IEEE conference on computer vision and pattern recognition},
  pages={4040--4048},
  year={2016}
}

@inproceedings{dosovitskiy2015flownet,
  title={Flownet: Learning optical flow with convolutional networks},
  author={Dosovitskiy, Alexey and Fischer, Philipp and Ilg, Eddy and Hausser, Philip and Hazirbas, Caner and Golkov, Vladimir and Van Der Smagt, Patrick and Cremers, Daniel and Brox, Thomas},
  booktitle={IEEE international conference on computer vision},
  pages={2758--2766},
  year={2015}
}

@inproceedings{jiang2021gma,
  title={Learning to estimate hidden motions with global motion aggregation},
  author={Jiang, Shihao and Campbell, Dylan and Lu, Yao and Li, Hongdong and Hartley, Richard},
  booktitle={IEEE/CVF international conference on computer vision},
  pages={9772--9781},
  year={2021}
}

@inproceedings{schops2017multi,
  title={A multi-view stereo benchmark with high-resolution images and multi-camera videos},
  author={Schops, Thomas and Schonberger, Johannes L and Galliani, Silvano and Sattler, Torsten and Schindler, Konrad and Pollefeys, Marc and Geiger, Andreas},
  booktitle={IEEE conference on computer vision and pattern recognition},
  pages={3260--3269},
  year={2017}
}

@article{bao2020instereo2k,
  title={Instereo2k: a large real dataset for stereo matching in indoor scenes},
  author={Bao, Wei and Wang, Wei and Xu, Yuhua and Guo, Yulan and Hong, Siyu and Zhang, Xiaohu},
  journal={Science China Information Sciences},
  volume={63},
  number={11},
  pages={212101},
  year={2020},
  publisher={Springer}
}

@inproceedings{scharstein2014high,
  title={High-resolution stereo datasets with subpixel-accurate ground truth},
  author={Scharstein, Daniel and Hirschm{\"u}ller, Heiko and Kitajima, York and Krathwohl, Greg and Ne{\v{s}}i{\'c}, Nera and Wang, Xi and Westling, Porter},
  booktitle={German conference on pattern recognition},
  pages={31--42},
  year={2014},
  organization={Springer}
}

\clearpage
\appendix

\setcounter{figure}{0}
\renewcommand{\thefigure}{A\arabic{figure}} 

\section{Supplementary Material}

\subsection{Need for Dense Cross-Spectral Correspondence}
\begin{figure}[h]
  \centering
    \includegraphics[width=0.45\linewidth]{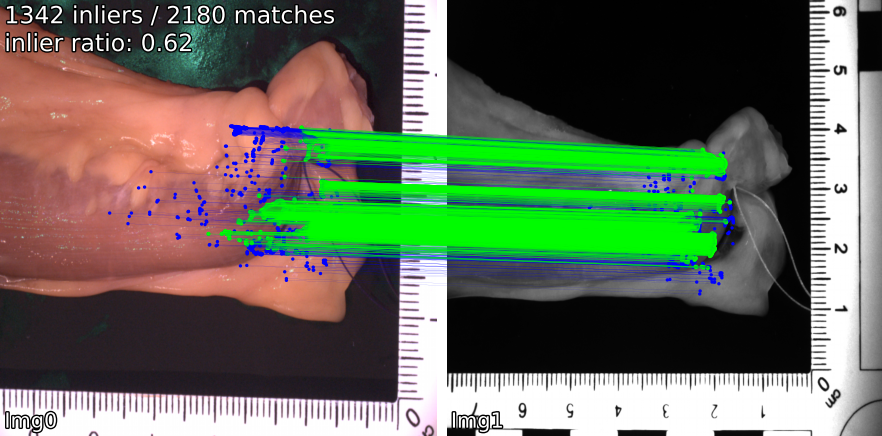}
    \includegraphics[width=0.45\linewidth]{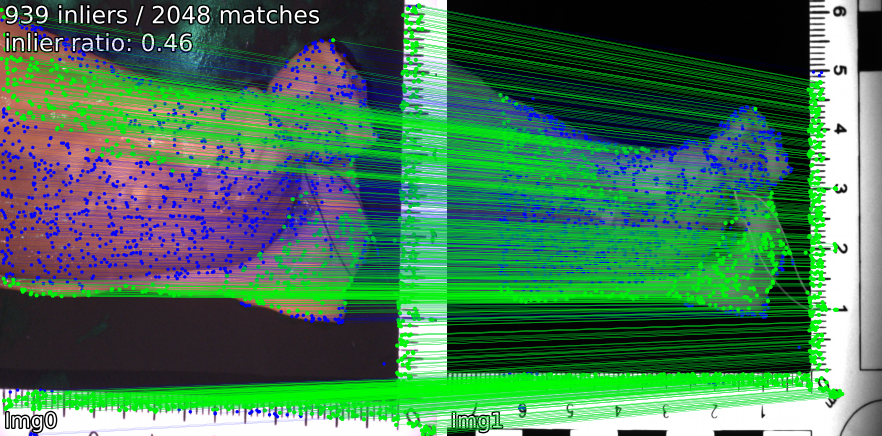}
    \caption{Representative correspondence results for an RGB-SWIR image pair obtained with DUSt3R (left) and RoMa (right). Reliable matches are concentrated in locally distinctive image regions, while substantial parts of the scene remain without correspondence support.}
    \label{fig:sota}
\end{figure}

\Cref{fig:sota} illustrates representative matching results obtained with DUSt3R \cite{wang2024dust3r} and RoMa \cite{edstedt2024roma} for images acquired with our RGB-SWIR setup. Under a reasonable confidence filtering, reliable matches are concentrated in a limited set of regions. Although both methods recover plausible correspondences in locally distinctive regions, where the intensity distribution shows similar behavior in both view, the resulting matches are spatially uneven and large parts of the scene remain without reliable correspondence support. Such locally concentrated matches may be sufficient for coarse geometric alignment or pose estimation, but they do not provide the complete pixel-wise mapping required for cross-spectral data fusion. In stereo-HSI and related multi-sensor imaging systems, spectral measurements must be transferred into a common spatial reference across the entire field of view to enable spatially coherent fusion and quantitative downstream analysis. A robust dense displacement field is therefore required not only at salient structures, but also across weakly textured and spectrally dissimilar regions where conventional feature matching frequently provides little or no coverage.

\subsection{Synth: Controlled Benchmark for Spectral Mismatch}
\begin{figure}[h]
  \centering
  	\includegraphics[width=.31\columnwidth]{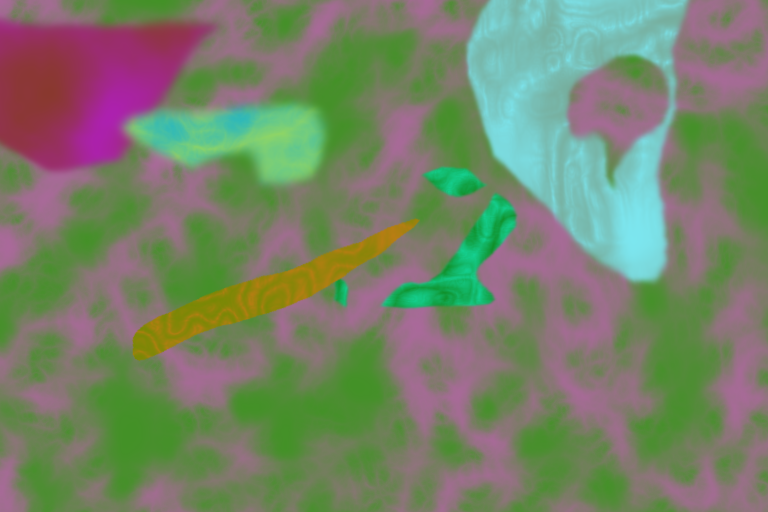}
	\includegraphics[width=.31\columnwidth]{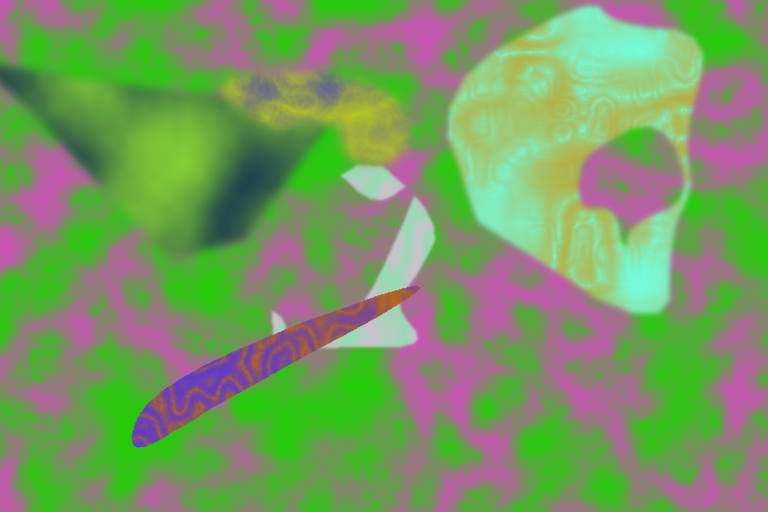}
	\includegraphics[width=.31\columnwidth]{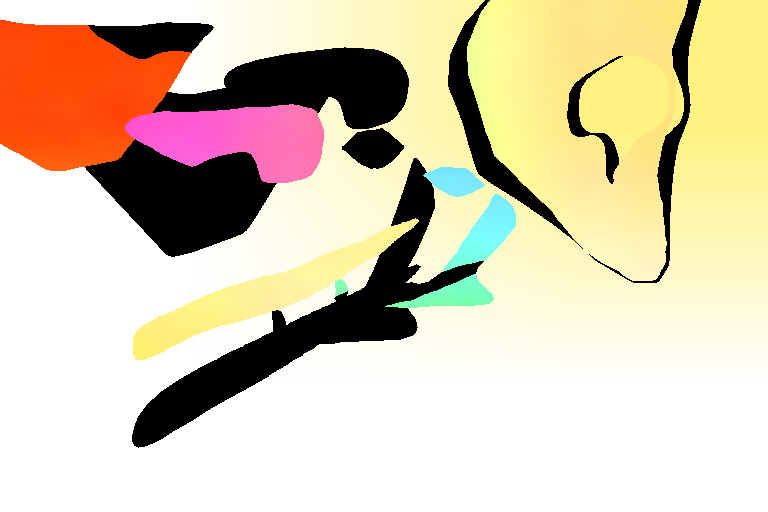}
  \caption{Examples of a generated triplet of the own created synthetic cross-spectral dense correspondence dataset (\textit{Synth-Extreme}). Left column: image $I_L$ Middle column: image $I_R$ Right column: displacement field $\mathbf{u}'$ of the scene.}
  \label{fig:synth}
\end{figure}

As described in the paper, \textit{Synth} is a synthetic generated dataset containing dense ground-truth displacement and spectral pairing metadata. We use measured reflectance spectra from the USGS Spectral Library \cite{kokaly2017usgs}, which is a curated reference collection of measured spectral reflectance signatures. It contains spectra of several thousand measured spectra of natural, biological, and man-made materials. The spectra ranges over a very broad wavelength range, from about $0.2-200 \mu$m, covering ultraviolet (UV), visual (VIS), near-infrared (NIR), shortwave infrared (SWIR), midwave infrared (MWIR), and longwave infrared (LWIR). A RGB rendered visualization of a randomly selected example is shown in \cref{fig:synth}.

\subsection{Data Augmentation by Spectral-Response Modulation}
In the second augmentation strategy (`Grayscale with spectral response modulation', cf.~\cref{sec:dataaugment}), the image pairs are converted to a single-channel representation either by channel-wise normalization followed by grayscale conversion or by randomly selecting one RGB channel independently for each view. One of the two resulting intensity images is additionally modified by a structured nonlinear radiometric transformation. This emulates sensor- and wavelength-dependent appearance changes and forces the networks to learn correspondences that are less dependent on direct photometric similarity. The transformations are sampled from a fixed family 18 functions:
\begin{itemize}
    \item identity and inversion mappings, $f(x)=x$ and $f(x)=1-x$,
    \item square-root variants, e.g., $f(x)=\sqrt{x}$, $f(x)=1-\sqrt{x}$, $f(x)=\sqrt{1-x}$, and $f(x)=1-\sqrt{1-x}$,
    \item power-law mappings, e.g., $f(x)=x^n$, $f(x)=1-x^n$, $f(x)=(1-x)^n$, and $f(x)=1-(1-x)^n$ for $n\in\{2,4\}$,
    \item logarithmic variants, e.g., $f(x)=\log_2(x+1)$, $f(x)=\log_2(2-x)$, $f(x)=1-\log_2(x+1)$, and $f(x)=1-\log_2(2-x)$.
\end{itemize}
These transformations preserve the geometric image structure but vary the radiometric relation between the two views in a controlled manner. In this way, cross-spectral image pairs are created that explicitly violate brightness constancy while remaining fully compatible with the original displacement annotations.

\begin{figure}[t]
  \centering
  \includegraphics[width=0.225\columnwidth]{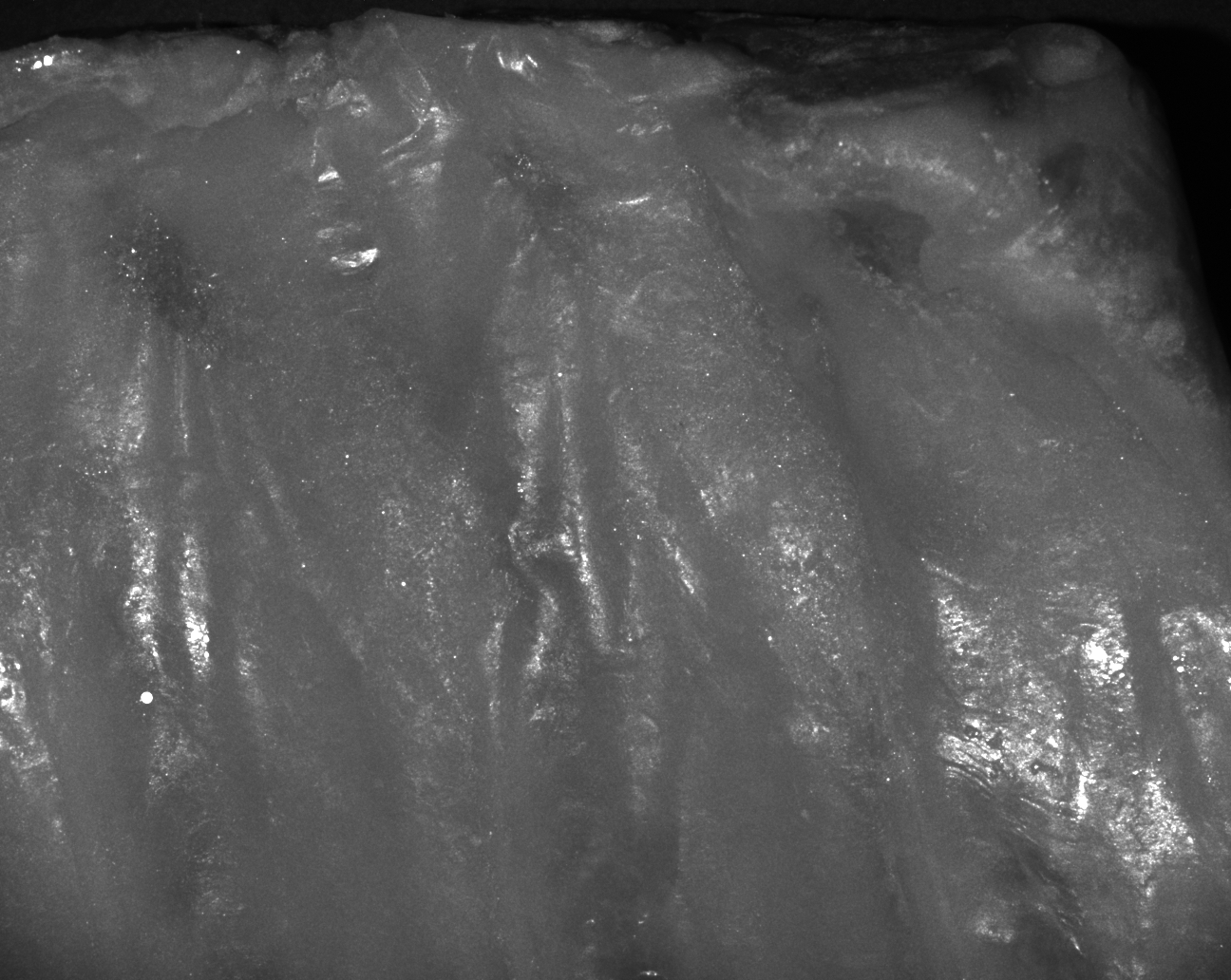}
  \includegraphics[width=0.239\columnwidth]{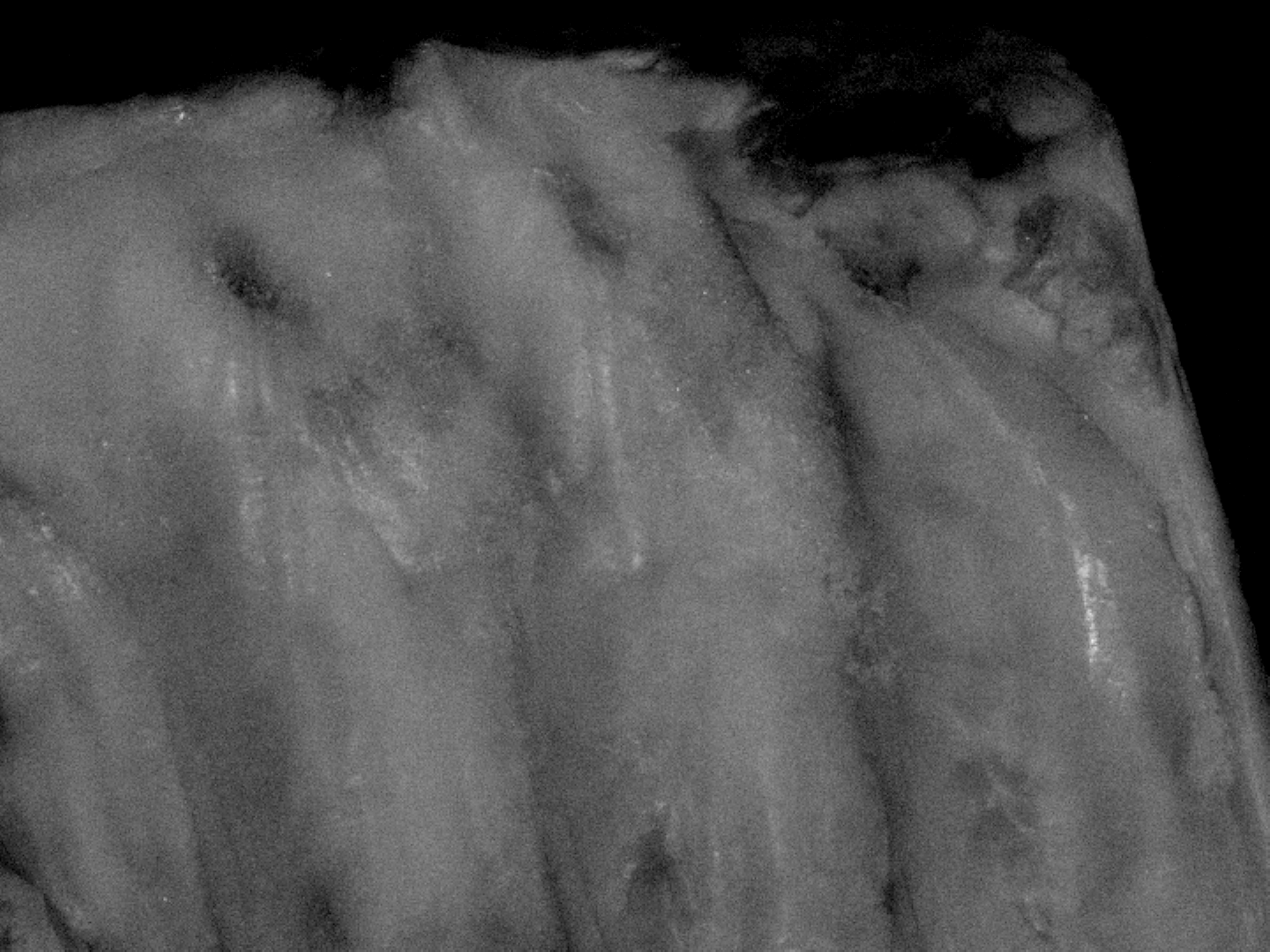}
  \includegraphics[width=0.225\columnwidth]{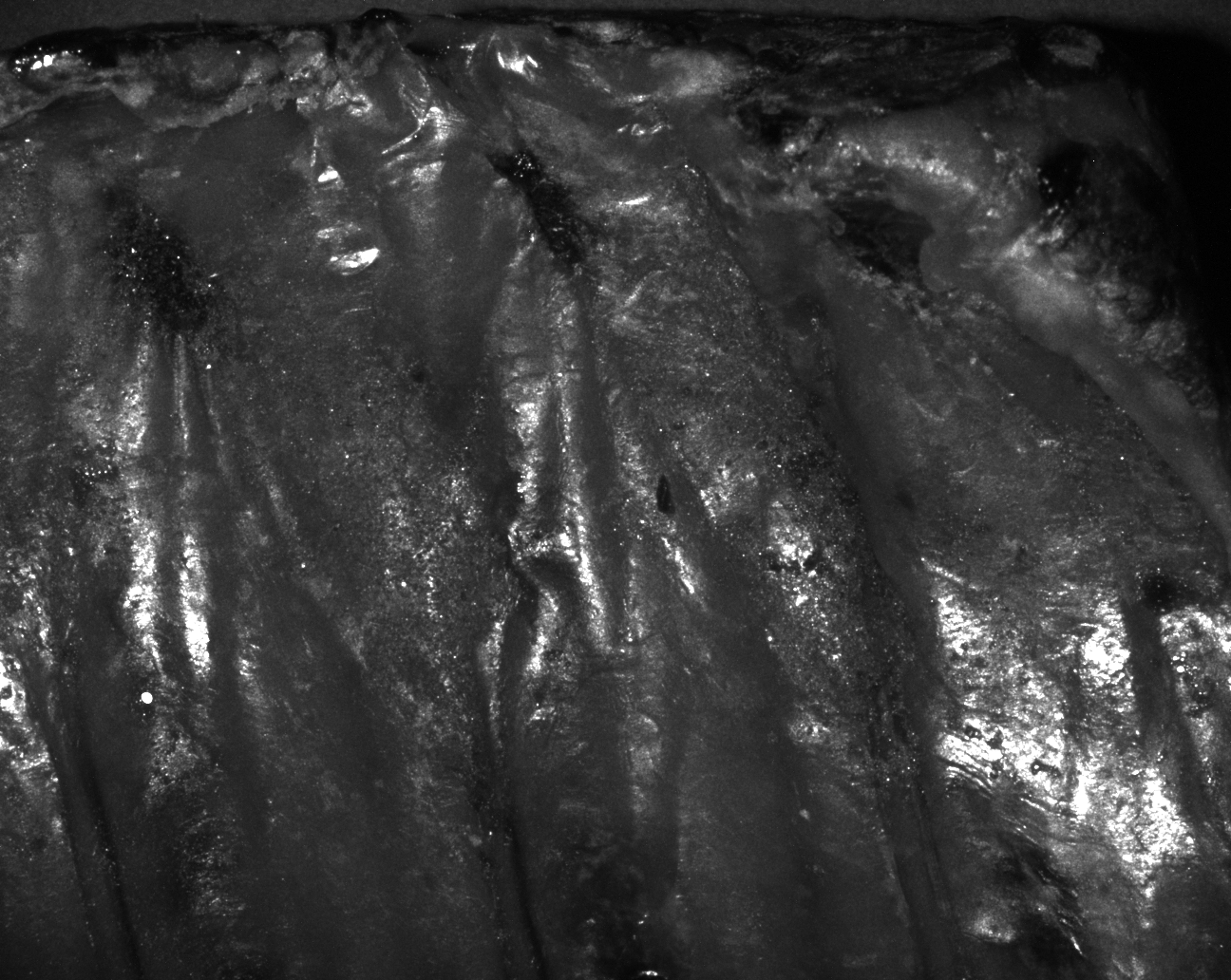}
  \includegraphics[width=0.239\columnwidth]{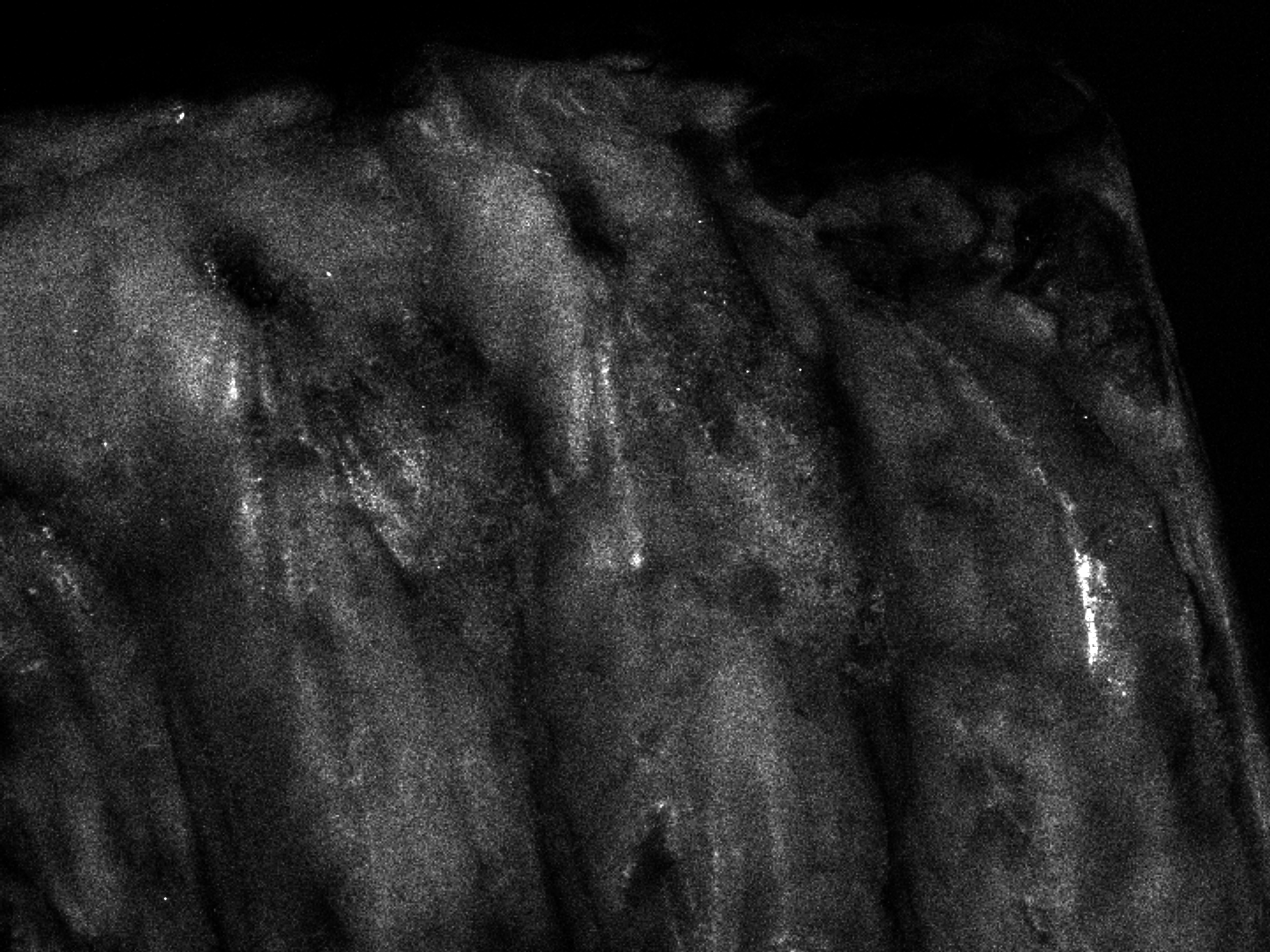}
   \caption{Examples of real RGB-SWIR radiometric relationships that are approximated by our functions. $1^{st}$ and $3^{rd}$ images show original SWIR image, while the $2^{nd}$ and $4^{th}$ images are the corresponding approximation using red and blue channel of the RGB capture, respectively.}
   \label{fig:swir}
\end{figure}

This function family has not been designed as a camera-specific physical sensor model. Rather, it is a compact approximation of the dominant radiometric effects that break brightness constancy. These effects include nonlinear contrast compression, contrast expansion and inversion. In order to verify that these transformations are not arbitrary, a comparison is made between real cross-spectral responses on RGB-SWIR data and the applied function family. For example, a $\lambda = 1064$\,nm response exhibited a contrast relationship similar to the square root of the red color channel, while a $\lambda = 1370$\,nm response resembled a higher-order nonlinear blue-channel response such as to the power of $n=4$, cf.~\cref{fig:swir}. This demonstrates that the selected transformations effectively capture realistic cross-spectral intensity relationships.

\subsection{Rationale and Limitations of the Single-Channel Input Representation} \label{supp:singlechannel}
A central design objective of our framework is its applicability across heterogeneous multi- and hyperspectral acquisition systems without adapting the correspondence architecture to a particular sensor configuration. Such systems can differ substantially in the number of acquired bands, their center wavelengths and bandwidths, as well as their covered spectral ranges. Directly using the complete spectral data cube as network input would therefore introduce sensor-specific channel semantics and potentially require architecture or training adaptations whenever the spectral configuration changes. We instead decouple geometric correspondence estimation from the dimensionality and wavelength configuration of the underlying spectral acquisition by representing each view through a normalized single-channel intensity image. Consequently, the same correspondence backbone can be applied to RGB, MSI, HSI, or single-band data while requiring only a common single-channel input interface.

Importantly, the single-channel representation does not necessarily imply averaging all available spectral bands. After band-wise normalization, the intensity representation can either be obtained by aggregating several bands or by directly selecting an individual spectral band. The latter is particularly relevant when spatial structures or boundaries are pronounced only within a specific wavelength range, as these features would be retained completely when the corresponding band is used as input. During training, individual normalized channels can additionally be sampled independently between the two views, preventing the correspondence model from relying on fixed wavelength or channel semantics and exposing it to wavelength-dependent appearance changes.

The single-channel formulation nevertheless represents a deliberate trade-off. An aggregation across bands may attenuate structures that are visible only within a narrow spectral range, whereas selecting an individual band does not allow the correspondence network to simultaneously exploit complementary structural information distributed across multiple wavelengths. Thus, a single forward pass cannot make full joint use of all spectral information contained in a multi- or hyperspectral acquisition. Adaptive spectral projections, learned wavelength selection, or spectral attention mechanisms could potentially exploit such complementary information more effectively. However, these approaches would generally introduce additional assumptions about the number, ordering, or spectral characteristics of the available bands and thereby reduce the sensor-independent and plug-and-play character of the proposed framework.

It is also important to distinguish the input representation used for correspondence estimation from the spectral information retained by the overall imaging pipeline. The single-channel projection is used only to estimate the geometric displacement field. The original multi-band measurements remain available and can subsequently be aligned using the estimated dense correspondences for spatially coherent spectral fusion and downstream analysis. Hence, the proposed representation does not discard spectral information from the acquisition or fusion pipeline; rather, it deliberately separates the task of estimating geometric correspondence from the subsequent exploitation of the full spectral information. This separation enables established dense correspondence architectures to operate across heterogeneous spectral systems without introducing sensor-specific network interfaces.

\subsection{t-SNE Visualization} \label{supp:tsne}
\begin{figure}[th]
  \centering
    \begin{subfigure}{0.49\linewidth} \centering
    \includegraphics[width=0.72\columnwidth]{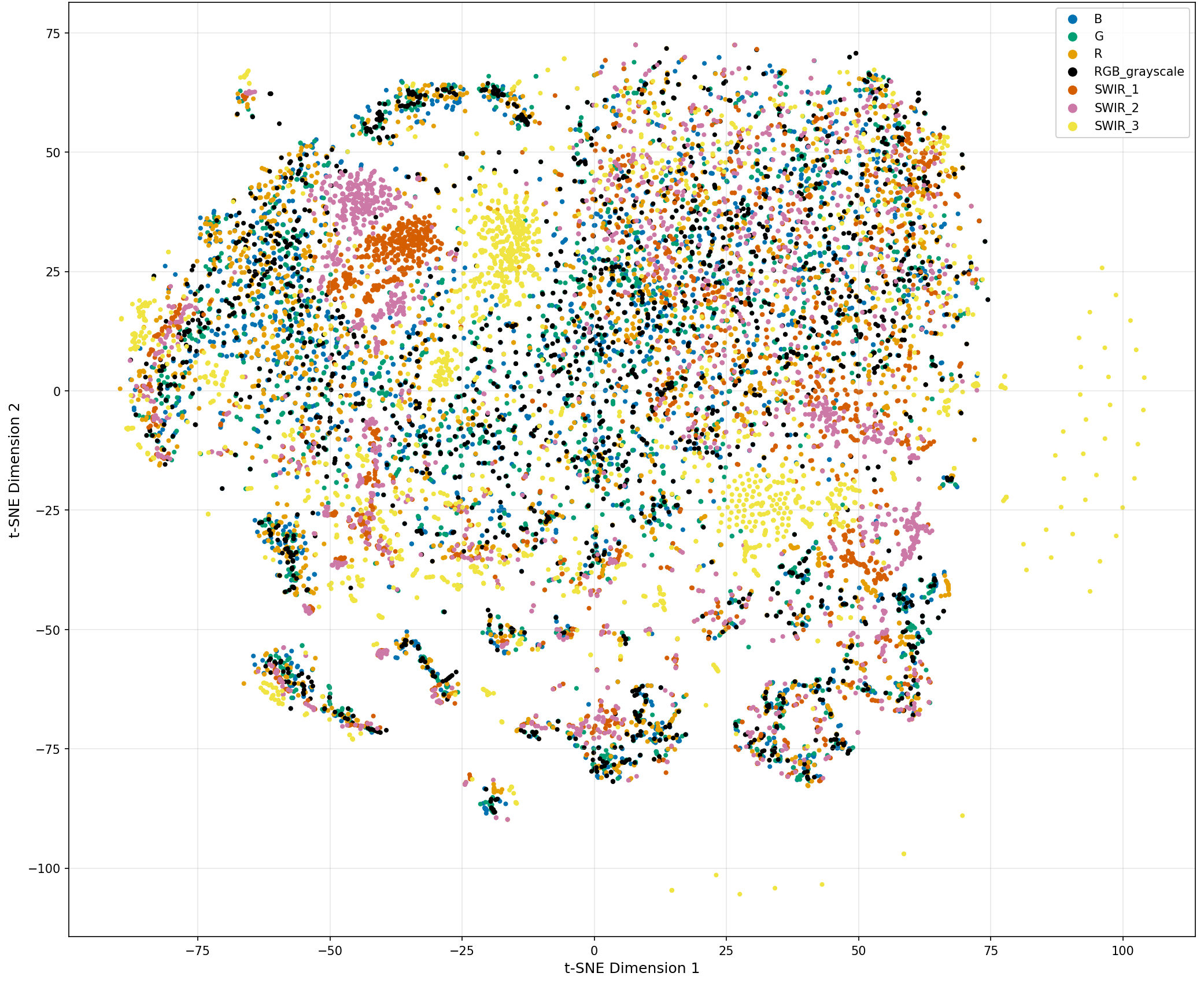}
    \caption{Original DIP} \label{fig:tsneA}
  \end{subfigure}
  \hfill
  \begin{subfigure}{0.49\linewidth} \centering
    \includegraphics[width=0.72\columnwidth]{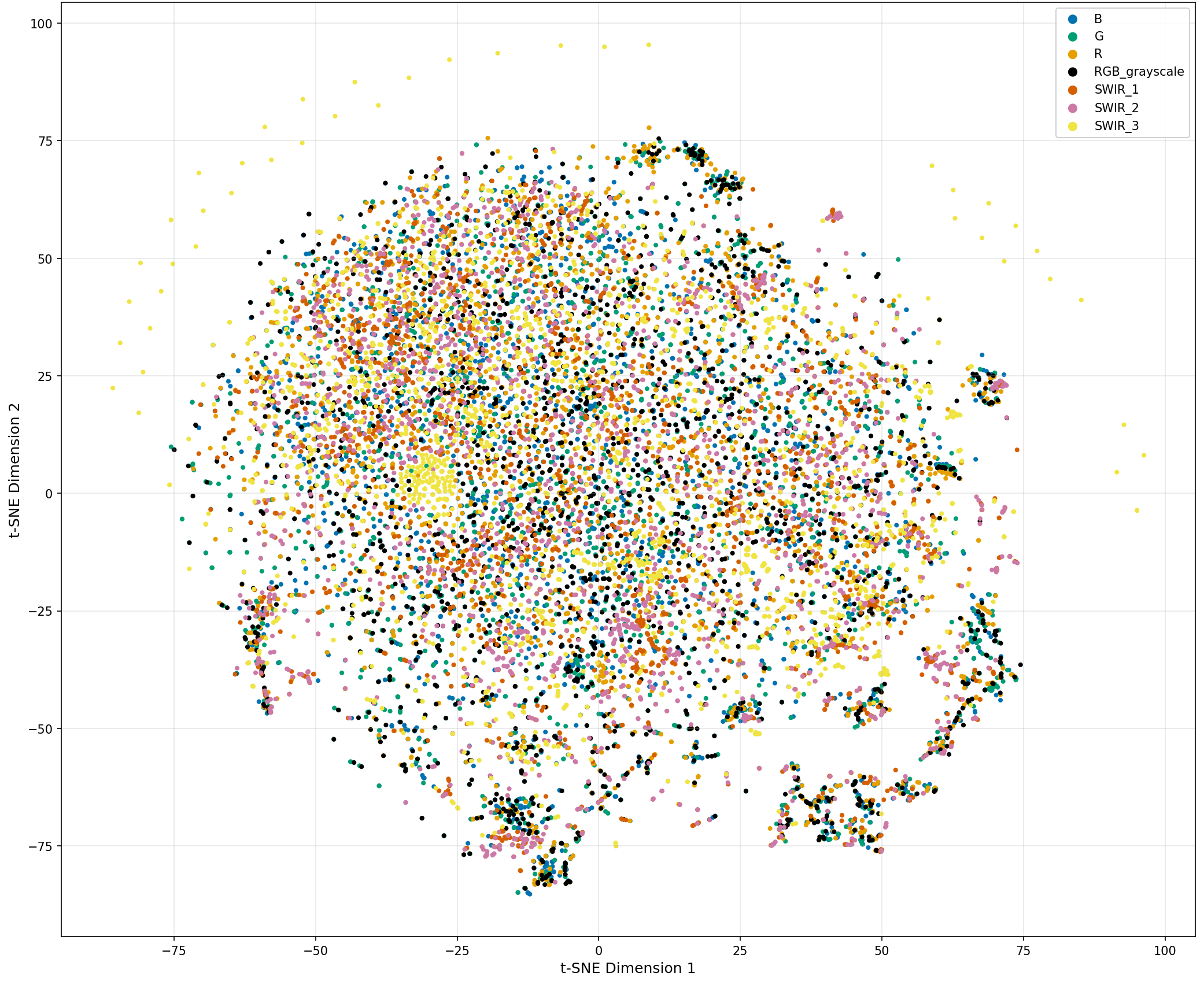}
    \caption{Cross-Spectral DIP} \label{fig:tsneB}
  \end{subfigure}
  \caption{A t-SNE visualization of a randomly chosen scene from the RGB-SWIR dataset. While the features show different clustering per view in the original model (a), all features are highly similar over different spectral modalities for the cross-spectral model (b) enabling better correspondence learning.}
  \label{fig:tsne}
\end{figure}

An additional qualitative evidence for the improved features in the cross-spectral backbones depicts t-SNE visualization, as it illustrates the shift of the backbone features.
An RGB-SWIR biomedical example is presented in \cref{fig:tsne}. 
For the original DIP model (\cref{fig:tsneA}), embeddings form distinct clusters depending on the spectral input variant, indicating that the representation remains modality-dependent even when the underlying scene content is identical. In contrast, the cross-spectral model (\cref{fig:tsneB}) produces embeddings that largely overlap across variants, suggesting that the learned features are dominated by structural cues and spatial content rather than sensor-specific radiometry. This reduction of modality-induced clustering is consistent with the higher feature correlations in \cref{tab:processing_adaptation} and with the improved correspondence accuracy on cross-spectral benchmarks.

\subsection{Qualitative Results} \label{supp:vis}
\begin{figure}[h]
  \centering
    \begin{subfigure}{0.49\linewidth} \centering
    \includegraphics[width=0.9\columnwidth]{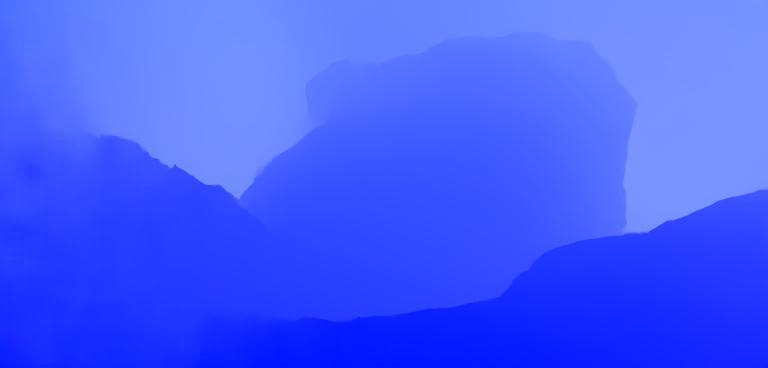}
    \caption{Original DIP}
  \end{subfigure}
  \hfill
  \begin{subfigure}{0.49\linewidth} \centering
    \includegraphics[width=0.9\columnwidth]{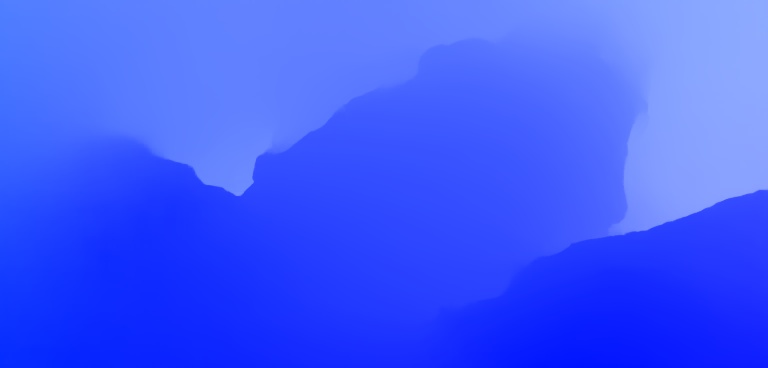}
    \caption{Cross-Spectral DIP}
  \end{subfigure}
    \begin{subfigure}{0.49\linewidth} \centering
    \includegraphics[width=0.9\columnwidth]{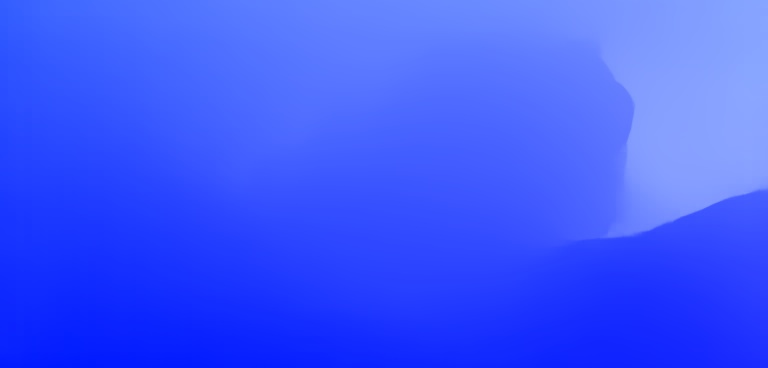}
    \caption{Original GMA}
  \end{subfigure}
  \hfill
  \begin{subfigure}{0.49\linewidth} \centering
    \includegraphics[width=0.9\columnwidth]{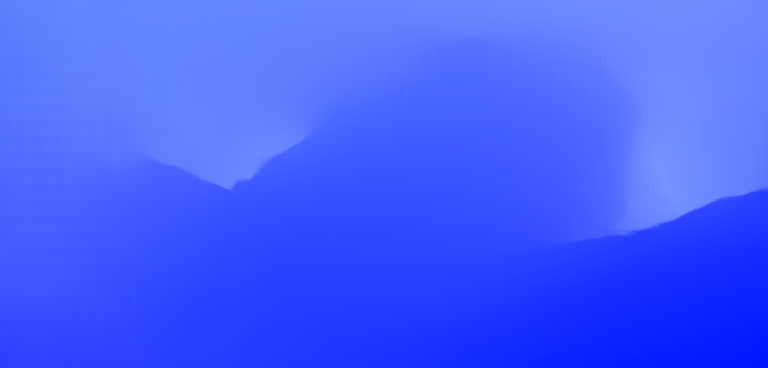}
    \caption{Cross-Spectral GMA}
  \end{subfigure}
    \begin{subfigure}{0.49\linewidth} \centering
    \includegraphics[width=0.9\columnwidth]{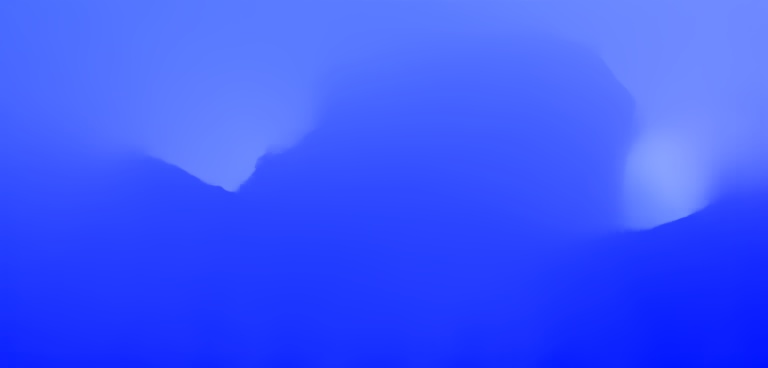}
    \caption{Original RAFT}
  \end{subfigure}
  \hfill
  \begin{subfigure}{0.49\linewidth} \centering
    \includegraphics[width=0.9\columnwidth]{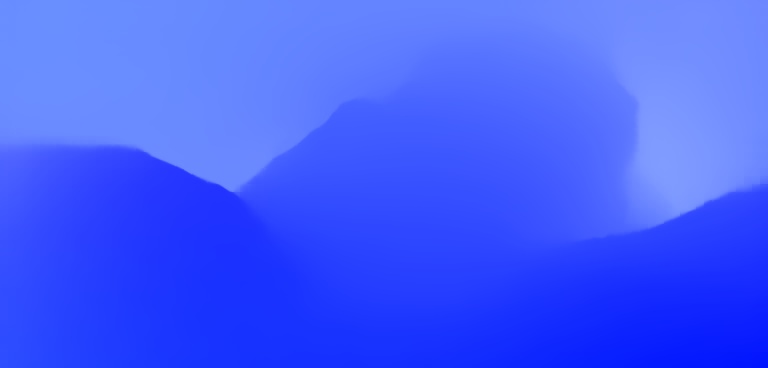}
    \caption{Cross-Spectral RAFT}
  \end{subfigure}
  \begin{subfigure}{0.49\linewidth} \centering
    \includegraphics[width=0.9\columnwidth]{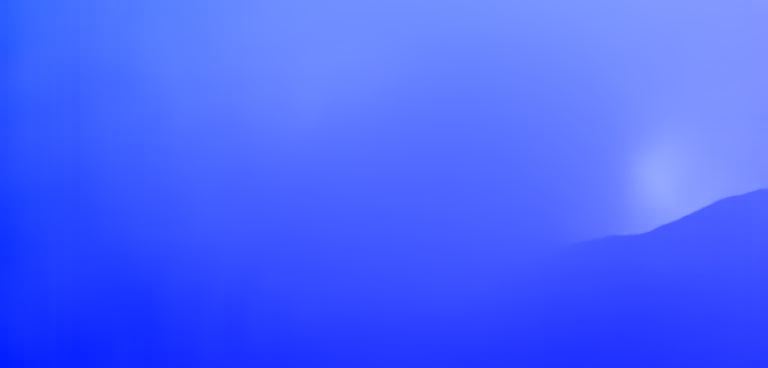}
    \caption{Original SKFlow}
  \end{subfigure}
  \hfill
  \begin{subfigure}{0.49\linewidth} \centering
    \includegraphics[width=0.9\columnwidth]{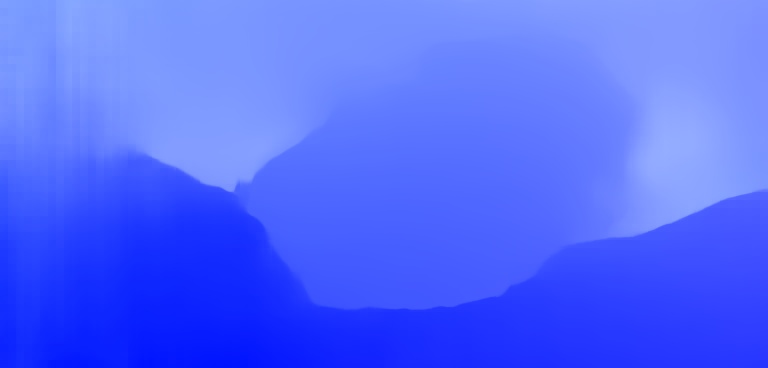}
    \caption{Cross-Spectral SKFlow}
  \end{subfigure}
  \caption{The qualitative results, i.e., displacement maps of the MSI stereo setup of the other utilized architectures of the same intraoperative scene as in \cref{fig:realdata}.}
  \label{fig:displacementMSI}
\end{figure}

\begin{figure}[H]
  \centering
    \begin{subfigure}{0.49\linewidth} \centering
    \includegraphics[width=0.9\columnwidth]{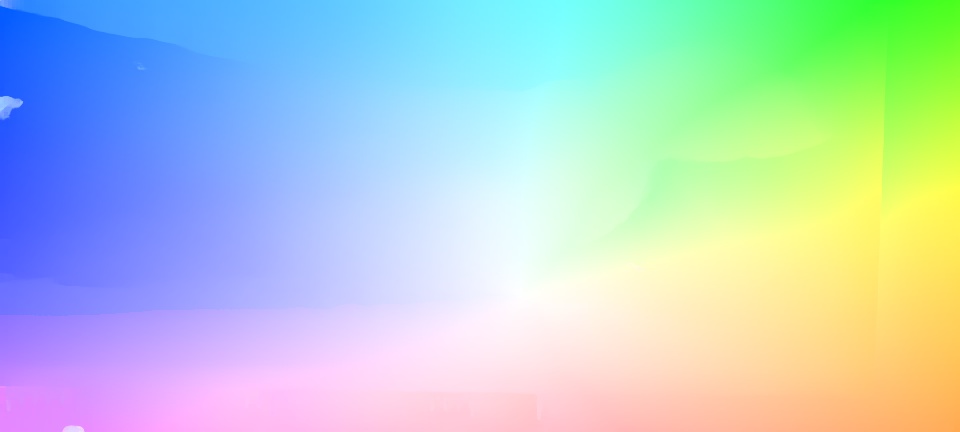}
    \caption{Original DIP}
  \end{subfigure}
  \hfill
  \begin{subfigure}{0.49\linewidth} \centering
    \includegraphics[width=0.9\columnwidth]{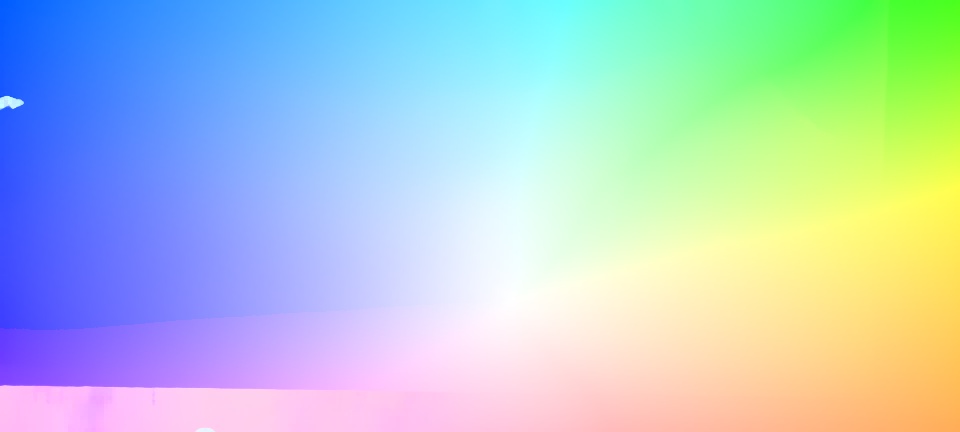}
    \caption{Cross-Spectral DIP}
  \end{subfigure}
    \begin{subfigure}{0.49\linewidth} \centering
    \includegraphics[width=0.9\columnwidth]{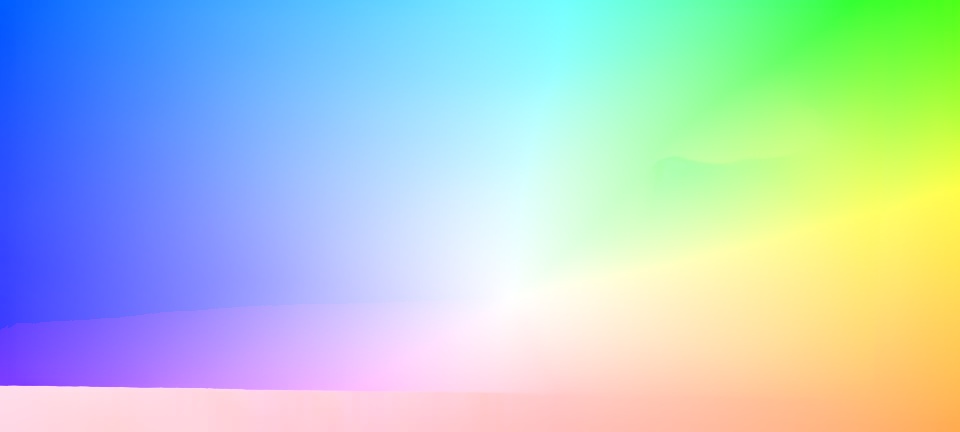}
    \caption{Original GMA}
  \end{subfigure}
  \hfill
  \begin{subfigure}{0.49\linewidth} \centering
    \includegraphics[width=0.9\columnwidth]{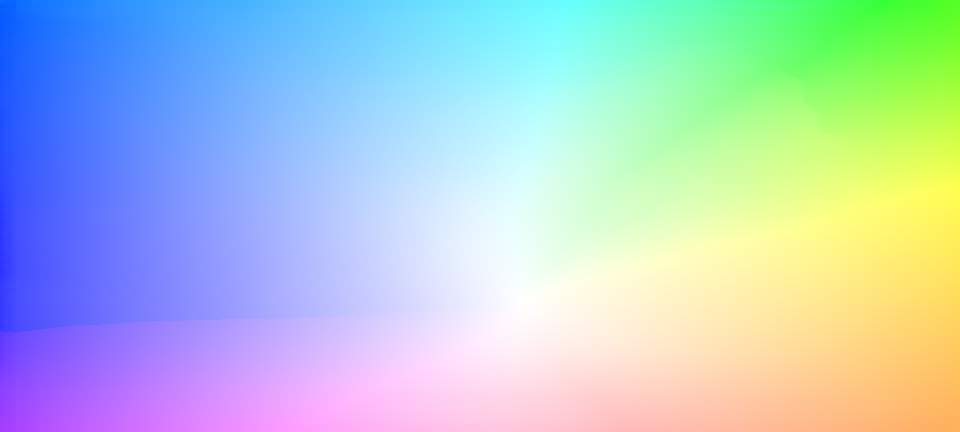}
    \caption{Cross-Spectral GMA}
  \end{subfigure}
    \begin{subfigure}{0.49\linewidth} \centering
    \includegraphics[width=0.9\columnwidth]{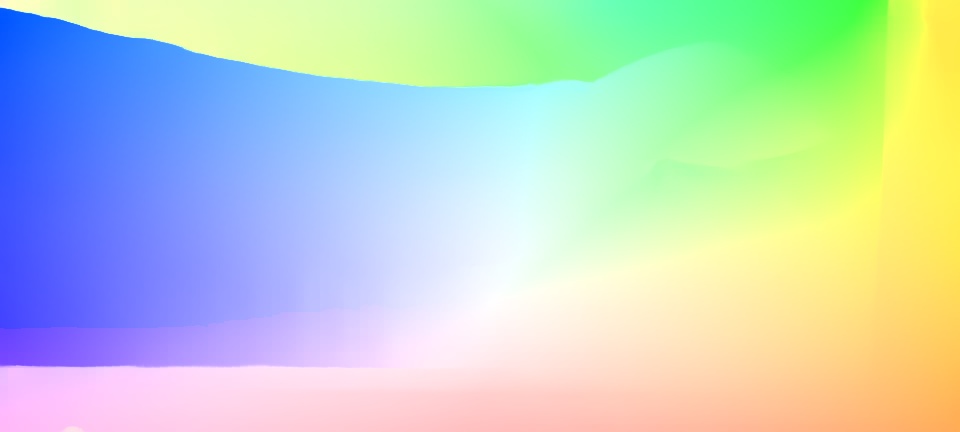}
    \caption{Original RAFT}
  \end{subfigure}
  \hfill
  \begin{subfigure}{0.49\linewidth} \centering
    \includegraphics[width=0.9\columnwidth]{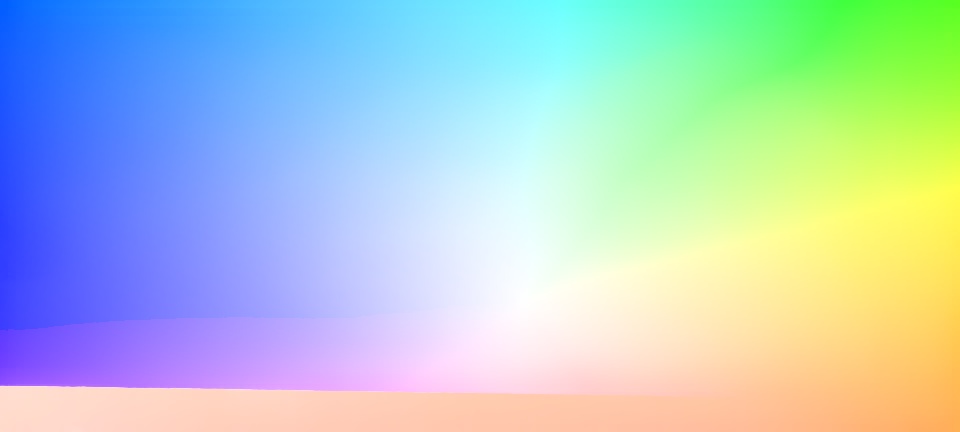}
    \caption{Cross-Spectral RAFT}
  \end{subfigure}
  \begin{subfigure}{0.49\linewidth} \centering
    \includegraphics[width=0.9\columnwidth]{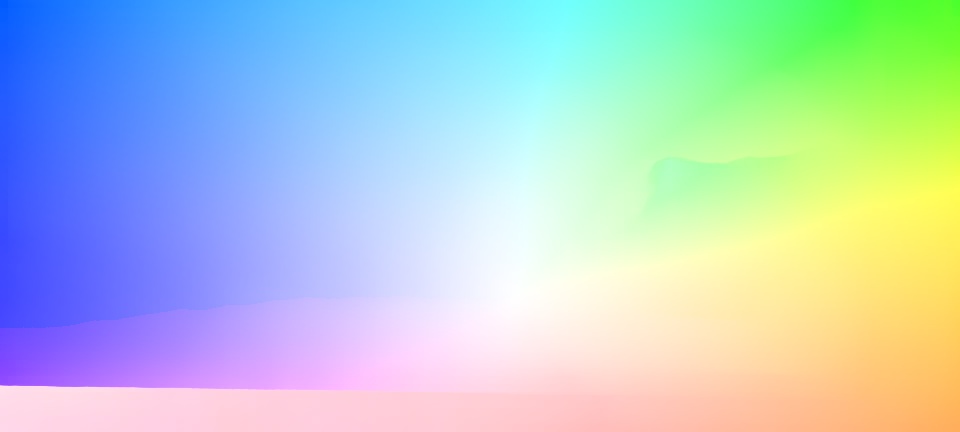}
    \caption{Original SKFlow}
  \end{subfigure}
  \hfill
  \begin{subfigure}{0.49\linewidth} \centering
    \includegraphics[width=0.9\columnwidth]{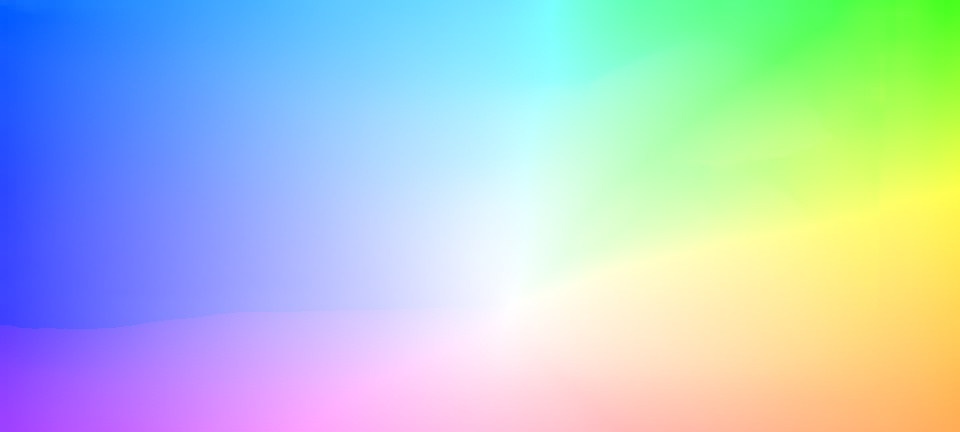}
    \caption{Cross-Spectral SKFlow}
  \end{subfigure}
  \caption{The qualitative results, i.e., displacement maps of the RGB-SWIR setup of the other utilized architectures of the same biomedical scene as in \cref{fig:realdata}.}
  \label{fig:displacementDD}
\end{figure}

\begin{figure}[h]
  \centering
    \begin{subfigure}{0.49\linewidth} \centering
    \includegraphics[width=0.9\columnwidth]{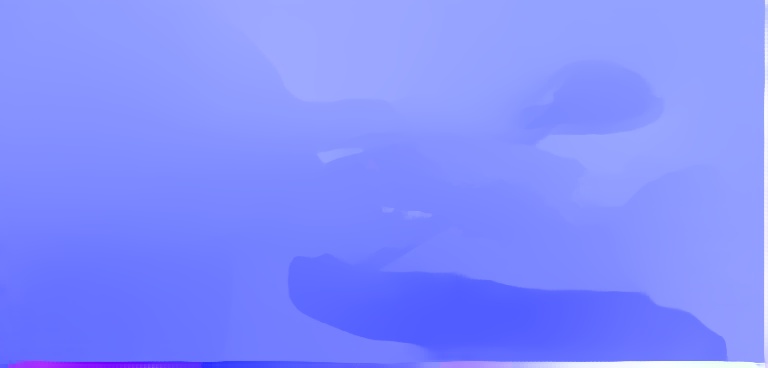}
    \caption{Original DIP}
  \end{subfigure}
  \hfill
  \begin{subfigure}{0.49\linewidth} \centering
    \includegraphics[width=0.9\columnwidth]{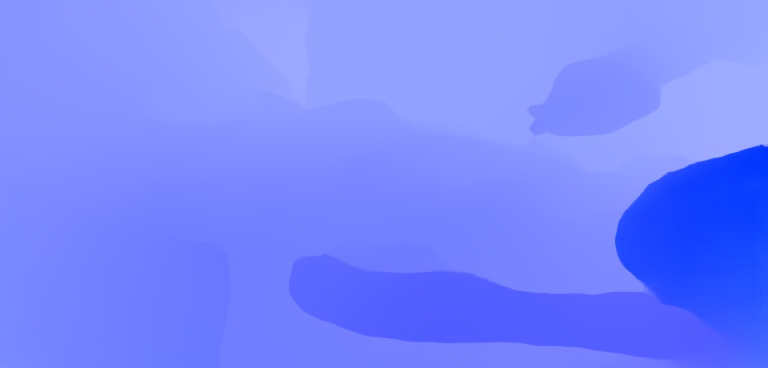}
    \caption{Cross-Spectral DIP}
  \end{subfigure}
    \begin{subfigure}{0.49\linewidth} \centering
    \includegraphics[width=0.9\columnwidth]{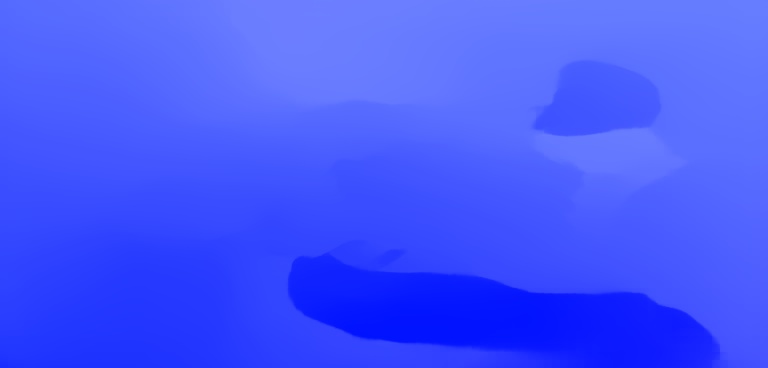}
    \caption{Original GMA}
  \end{subfigure}
  \hfill
  \begin{subfigure}{0.49\linewidth} \centering
    \includegraphics[width=0.9\columnwidth]{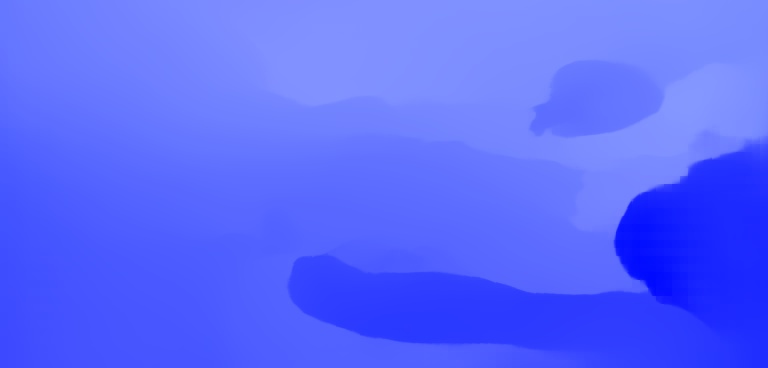}
    \caption{Cross-Spectral GMA}
  \end{subfigure}
    \begin{subfigure}{0.49\linewidth} \centering
    \includegraphics[width=0.9\columnwidth]{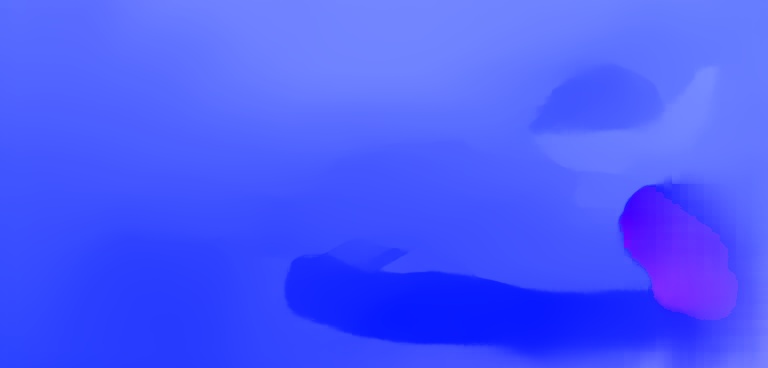}
    \caption{Original RAFT}
  \end{subfigure}
  \hfill
  \begin{subfigure}{0.49\linewidth} \centering
    \includegraphics[width=0.9\columnwidth]{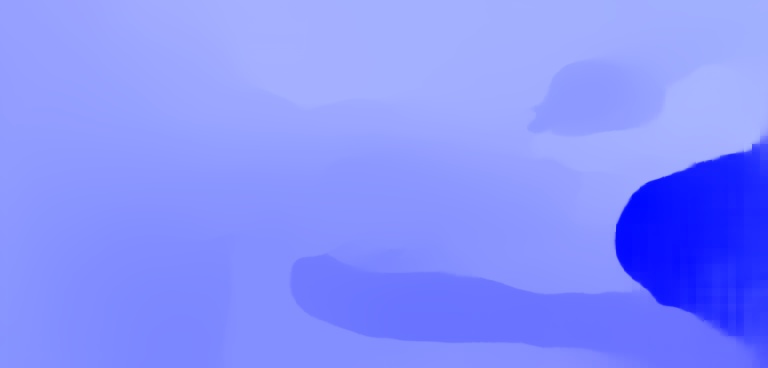}
    \caption{Cross-Spectral RAFT}
  \end{subfigure}
  \begin{subfigure}{0.49\linewidth} \centering
    \includegraphics[width=0.9\columnwidth]{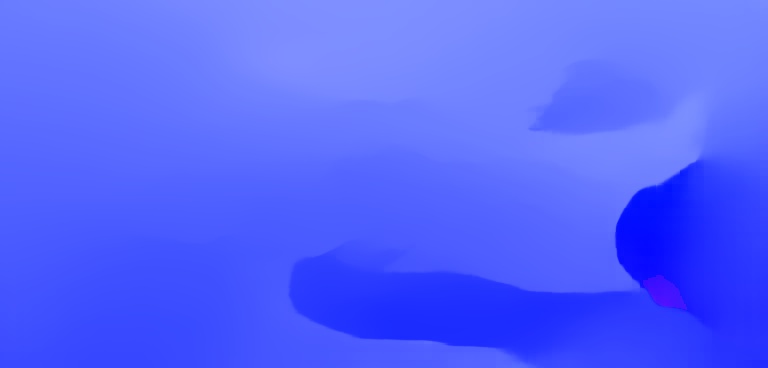}
    \caption{Original SKFlow}
  \end{subfigure}
  \hfill
  \begin{subfigure}{0.49\linewidth} \centering
    \includegraphics[width=0.9\columnwidth]{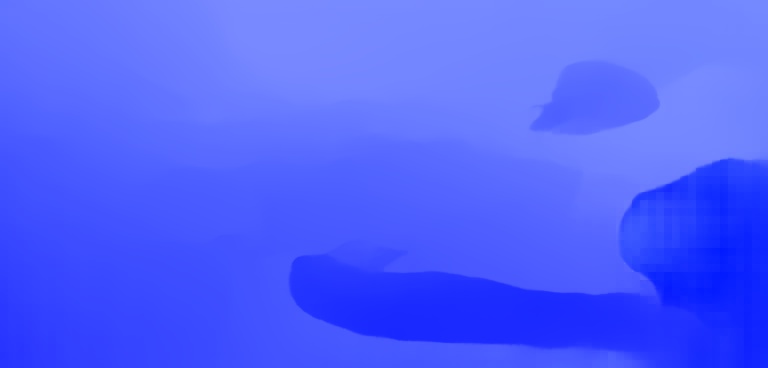}
    \caption{Cross-Spectral SKFlow}
  \end{subfigure}
  \caption{The qualitative results, i.e., displacement maps of the HSI light-field setup of the other utilized architectures of the same surgical scene as in \cref{fig:realdata}.}
  \label{fig:displacementLF}
\end{figure}

In the following \cref{fig:displacementMSI,fig:displacementDD,fig:displacementLF} the displacement maps of the other four analyzed architectures DIP, GMA, RAFT, and SKFlow (original model is shown in the left view, while the cross-spectral model is shown on the right) are presented for the same medical scenes that are used in \cref{fig:realdata}. It is visible, that all displacements maps of the cross-spectral models yield coherent and meaningful results. In contrast, the displacement maps from the original models exhibit artifacts, displacement flips, or missing content.

\end{document}